\pdfoutput=1

\documentclass[11pt]{article}

\usepackage[final]{acl}

\usepackage{subcaption}

\usepackage{times}
\usepackage{latexsym}

\usepackage[T1]{fontenc}

\usepackage[utf8]{inputenc}

\usepackage{microtype}

\usepackage{inconsolata}

\usepackage{graphicx}
\usepackage{xspace}
\usepackage{booktabs}
\newcommand{\ourmodel}{\textsc{DesignCLIP}\xspace}

\newcommand{\hhs}[1]{\textcolor{blue}{[Shomee: #1]}}
\newcommand{\todo}[1]{\textcolor{red}
{[TODO: #1]}}

\usepackage{amssymb}

\usepackage{multirow}
\usepackage{amsmath}
\definecolor{brickred}{RGB}{203, 65, 84}

\title{\ourmodel: Multimodal Learning with CLIP for Design Patent Understanding}

\author{%
  Zhu Wang \qquad Homaira Huda Shomee    \qquad Sathya N. Ravi \qquad Sourav Medya\\
  Department of Computer Science, University of Illinois Chicago\\
  \texttt{\{zwang260,hshome2,sathya,medya\}@uic.edu} \\
  }

\begin{document}
\maketitle
\begin{abstract}
In the field of design patent analysis, traditional tasks such as patent classification and patent image retrieval heavily depend on the image data. However, patent images---typically consisting of sketches with abstract and structural elements of an invention---often fall short in conveying comprehensive visual context and semantic information. This inadequacy can lead to ambiguities in evaluation during prior art searches. Recent advancements in vision-language models, such as CLIP, offer promising opportunities for more reliable and accurate AI-driven patent analysis. In this work, we leverage CLIP models to develop a unified framework \ourmodel for design patent applications with a large-scale dataset of U.S. design patents. 
To address the unique characteristics of patent data, \ourmodel incorporates class-aware classification and contrastive learning, utilizing generated detailed captions for patent images and multi-views image learning. We validate the effectiveness of \ourmodel across various downstream tasks, including patent classification and patent retrieval. Additionally, we explore multimodal patent retrieval, which provides potential to enhance creativity and innovation in design by offering more diverse sources of inspiration. Our experiments show that \ourmodel consistently outperforms baseline and SOTA models in patent domain on all tasks. Our findings underscore the promise of multimodal approaches in advancing patent analysis. The codebase is available here: \url{https://anonymous.4open.science/r/PATENTCLIP-4B3F/README.md}
\end{abstract}

\section{Introduction}
\label{sec:intro}

Patents are units for the state of the art innovation, designs and technological advancements as well as they offer legal protection for inventors' intellectual property \cite{moser2013patents}. Patents grant the owner the authority to prevent others from manufacturing, utilizing, or distributing the patented invention without their permission \footnote{\url{https://www.uspto.gov/patents/basics}}. Among the two popular types, \textit{utility patents} are granted for innovations in processes, machines, manufactures, or compositions of matter, including improvements, \textit{design patents} are awarded for new, original, and ornamental designs applied to manufactured items. While utility patents have been well studied \cite{fall2003automated,kamateri2022automated,siddharth2022enhancing,kang2020patent}, design patents remain relatively underexplored. 


One of the most important design patent tasks is patent retrieval. This aims to determine the novelty of the patent and prevent infringements. A patent is only granted if the design significantly differs from existing ones. Most previous research \cite{higuchi2023patent,lo2024large} focus primarily on image-to-image retrieval since there is a lack of informative text descriptions in design patents. Patent classification \cite{rademaker2000classification,kamateri2024will} is another task by patent reviewers who classify applications into various subject matters and assign design classification codes. In the US design patent system, there are 33 classes which also contain various subclasses. Automating the patent classification process can save an enormous amount of time.

\begin{figure*}
         \centering
         \includegraphics[width=.95\textwidth]{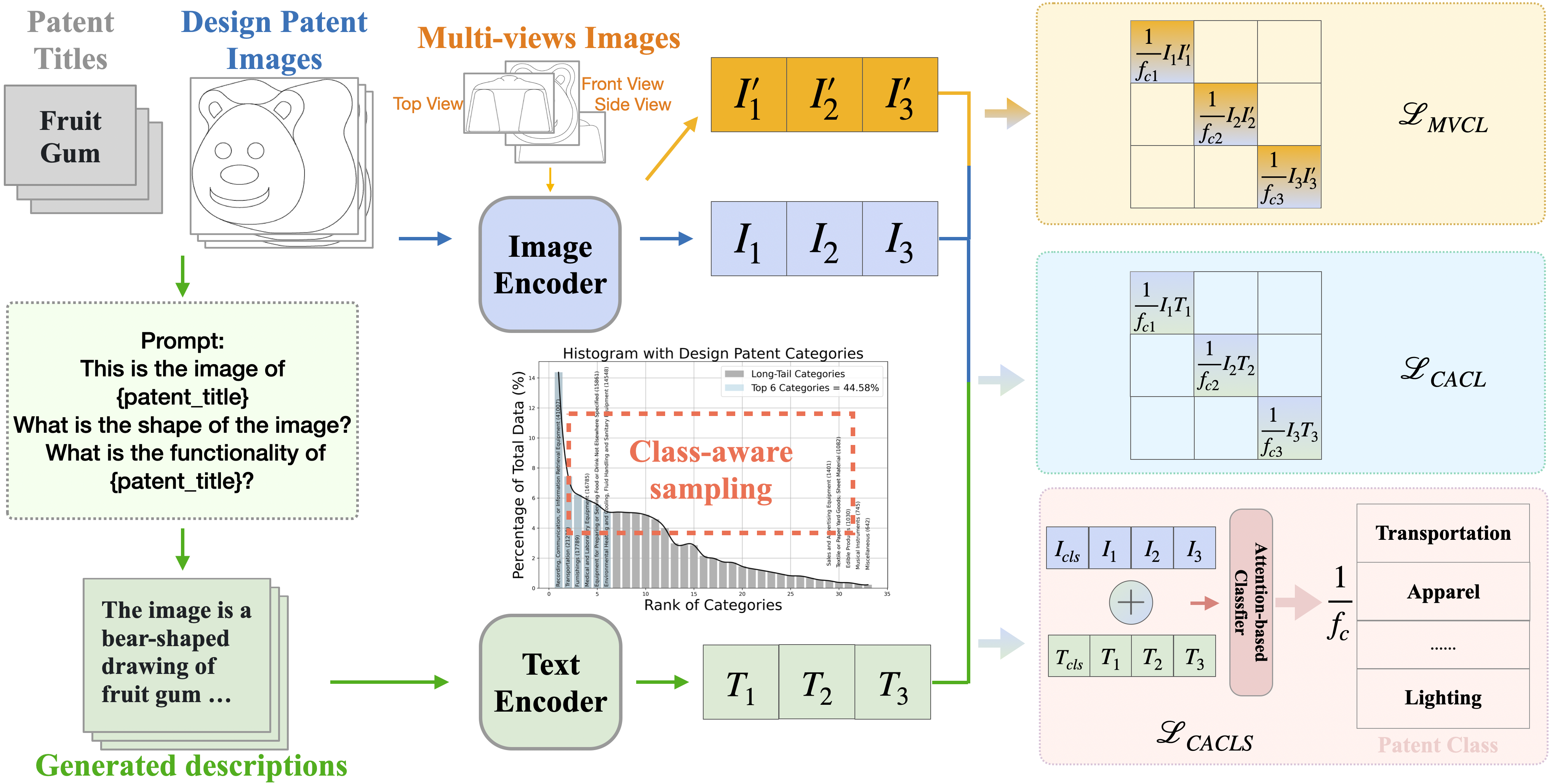}
                 
        \caption{The framework for \ourmodel. The inputs are patent images and simple text (title). First, we generate descriptions and pass them along with images to text and image encoders accordingly. Then, we pre-train CLIP-based models with proposed class-aware sampling and learning and multi-views image contrastive learning on 285,000 design patents. $\mathcal{L}_{\text{MVCL}}, \mathcal{L}_{\text{CACL}}$, and $\mathcal{L}_{\text{CACLS}}$ denote the multi-view contrastive loss (Eq. \ref{eq:mvcl}), class-aware contrastive loss (Eq. \ref{eq:mclip}), and class-aware classification loss (Eq. \ref{eq:cacls}) respectively. }
        \label{fig:pipeline}
\end{figure*}

Solving the above tasks would need sophisticated multimodal techniques due to several challenges. Design patent images are sketches that provide detailed design information and differ significantly from natural images (please see the examples of patent designs in Figure \ref{fig:pipeline}). Additionally, the class distribution in the design patent dataset is highly imbalanced, with the six most frequent classes accounting for almost half of the total data (Figure \ref{fig:distribution}). Furthermore, the aim is to build one single model that can perform the patent classification as well as retrieval.\textit{ There are currently no multimodal models that are designed being aware of these challenges and specifically tailored for design patents.}

\textbf{Our contributions.} In this paper, we develop a new CLIP-based framework to facilitate multimodal analyses on design patents. Our major contributions are as follows. 
\begin{itemize}
\item \textbf{Multimodal Analysis. }We address the problem of multimodal analysis in design patents by building a comprehensive AI-based tool. The goal is to address the time-consuming patent tasks such as patent classification and patent retrieval efficiently. 

\item \textbf{Our Proposed} \ourmodel: We modify CLIP \cite{clip} by incorporating the domain knowledge from design patents. It uses a class-aware learning method to balance the long-tailed distribution of classes in design patents, and is pre-trained with multiple tasks, including patent classification and multi-view image-image contrastive learning.

\item \textbf{Experiments}: We conduct comprehensive experiments showing that our methods outperforms baseline models and the state-of-art patent image retrieval models on all downstream tasks and gain notable improvements on design patent representations.

\end{itemize}

\section{Background on Patent Analysis}
The patent retrieval task focuses on efficiently retrieving relevant patent documents and images based on search queries. In design patents, this task is focused on image-to-image retrieval, where the objective is to find visually similar design images that match a given image query. \cite{Kucer_2022_WACV} implement various models such as ResNet50 \cite{he2016deep}, and Sketchy RN50 \cite{sangkloy2016sketchy}. Their patent-specific models are initially pre-trained on ImageNet \cite{deng2009imagenet} and fine-tuned on the DeepPatent dataset. Similarly, \cite{higuchi2023patent, higuchi2023patent2} use a deep metric learning framework, utilizing cross-entropy methods like InfoNCE with ArcFace. On the other hand, \cite{lo2024large} implement several advanced models, such as ViT \cite{dosovitskiy2020image} and Swin \cite{liu2021swin}, along with more recent multimodal  models such as BLIP-2 \cite{blip2}, and GPT-4V \cite{achiam2023gpt}. They introduce a novel approach by proposing a language-informed strategy for learning features from patent images. 
\textit{Notably, there is currently no specific machine learning or AI-based research focused on the critical constraints (e.g., classes and multi-views) from the design patents and we address this in this paper.} 

While utility patents are different from design patents, the images in utility patents are not well-studied. \cite{ghauri} classify utility patent images into distinct types of visualizations, including graphs, block circuits, flowcharts, and technical drawings. They employ the CLIP model \cite{clip}, integrated with a Multi-layer Perceptron and various Convolutional Neural Networks \cite{krizhevsky2012imagenet} architectures, to enhance the precision of patent image classification. IMPACT dataset \cite{shomee2024impact} is a comprehensive and large-scale resource comprising over 500,000 U.S. design patents issued between 2007 and 2022. It includes 3.61 million figures accompanied by detailed captions, titles, and metadata, offering a rich multimodal dataset that integrates visual and textual information. \textit{In this paper, we mainly focus on design patents using IMPACT dataset and address the two primary tasks associated with the design patents.} We demonstrate the design patent retrieval task in three different formats: text-to-image, image-to-text, and image-to-image. Additionally, the classification of design patent images into 33 subject matter (class) and various subclasses is a detailed and structured approach to organizing design patents. We address the classification task by categorizing patents into their class level categories using both text-based captions and images.

\begin{figure}[ht]
         \centering
         \includegraphics[width=.95\columnwidth]{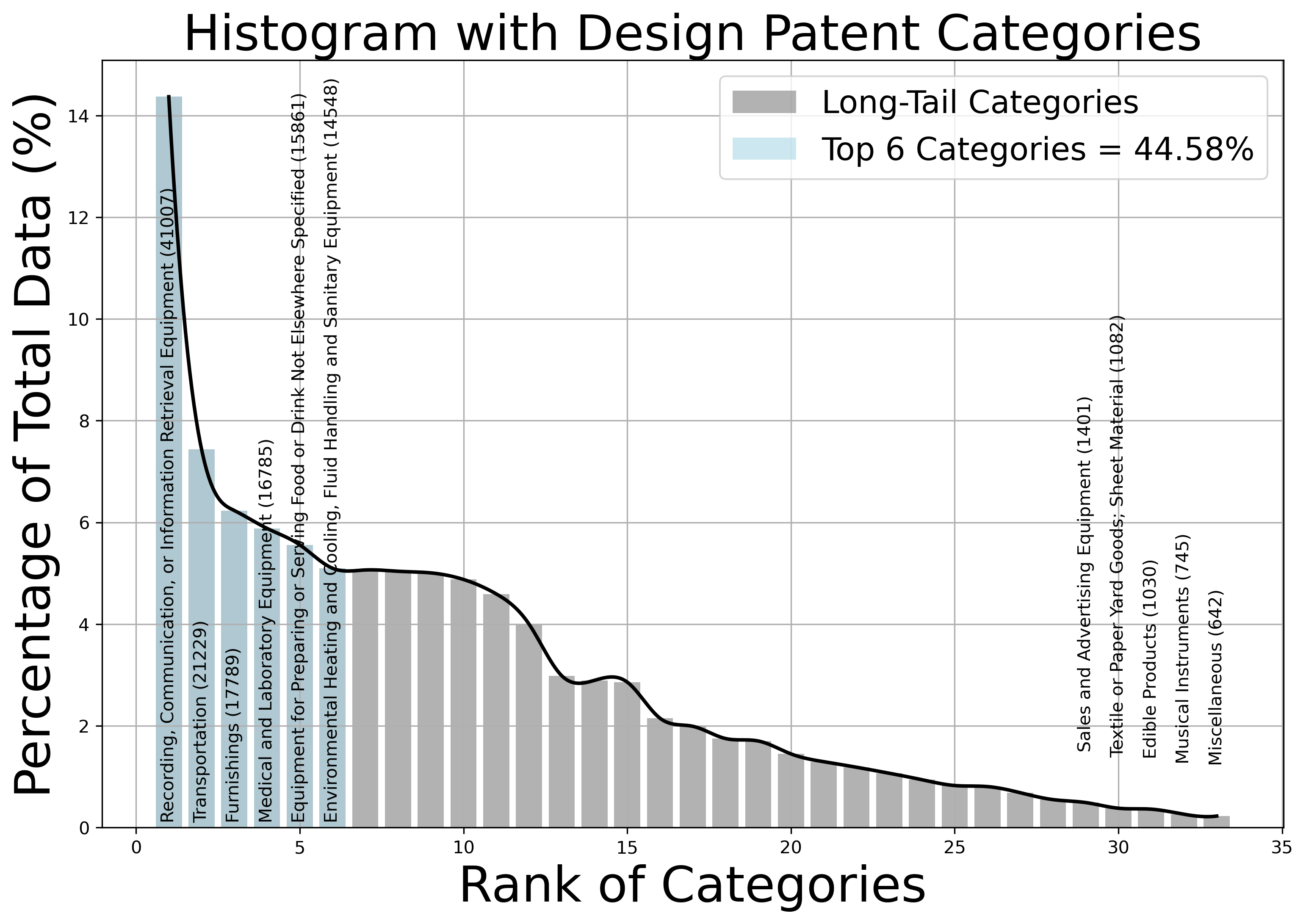}
        
         \label{fig:distribution}

        \caption{Design patent data category distributions in our test data. Top 6 categories consists of 44.58\% of all the data which shows a long tail distribution.}

        
        \label{fig:distribution}
\end{figure}

\section{\ourmodel}
In this work, we address the challenges of design patent analysis by adapting the CLIP vision-language model family, highlighting its simplicity and effectiveness for this domain specialized application. The major components of our proposed framework, \ourmodel, are as follows: \textit{First,} considering the long-tailed distribution among the classes of design patents, we begin with proposing a class-aware learning method to resample construct pairs in the batch. This helps our method \ourmodel to balance learning from all the classes. 
\textit{Then}, we pre-train \ourmodel with multiple tasks, including patent classification, image-caption contrastive loss, and multi-view image-image contrastive loss. Our architecture is illustrated in Figure \ref{fig:pipeline}.

\textbf{Brief Review of CLIP Framework:} CLIP \cite{clip}, a multimodal model uses contrastive learning to bridge the gap between text and images. At its foundation is the principle of acquiring perceptual understanding from the guidance of natural language. During its pre-training phase, CLIP learns to identify if a text snippet and an image are matched in its dataset. The training involves a series of five ResNets and three Vision Transformers to facilitate zero-shot classification. CLIP has been popular across various fields including medical imaging \cite{med1, med2, med3}, robotics \cite{robotics1, robotics2, robotics3}, biodiversity monitoring \cite{bio}, e-commerce \cite{ecommerce}, educational technology \cite{edu} and showed great performance.

The popular CLIP \cite{clip} is a pre-trained model to learn image-text pairs with contrastive loss. This approach enables the model to differentiate between semantically similar and dissimilar data points, enhancing its ability to capture meaningful relationships between visual and textual information. CLIP is widely applied for a variety of multimodal tasks, including zero-shot classification, and multimodal search. Specifically, given an image, CLIP can retrieve the most relevant text descriptions (e.g., captions or labels). Given image-text pairs $\{(v_i,t_i)\}^N_{i=1}$, the vanilla loss function in CLIP is defined as:

\begin{equation}
\mathcal{L}_{\text{CLIP}} = - \log \frac{\exp(\text{sim}(\mathbf{v}_i, \mathbf{t}_i)/\tau)}{\sum_{k=1}^{N} \exp(\text{sim}(\mathbf{v}_i, \mathbf{t}_k)/\tau)}, \label{eq:clip}
\end{equation}
where $\text{sim}(\cdot)$ is the cosine similarity between the image embedding $v_i$ and the text embedding $t_i$, and $\tau$ is a temperature parameter.

\subsection{Class-aware Learning} \label{sec:cal}
As illustrated in Figure \ref{fig:distribution}, the class (category) distribution in the design patent dataset is highly imbalanced, among which the top six most frequent classes constitute 44.58\% of the total training data. This imbalance presents a significant challenge for the standard CLIP contrastive learning method. This long-tailed recognition problem tends to be biased towards the head classes and leads to suboptimal performance on the less frequent classes. In \ourmodel, we address this problem by utilizing the marginal distribution in two ways: Class-aware Sampling and a modified Class-aware Contrastive Loss.

\subsubsection{Class-aware Sampling} While training \ourmodel, it is essential for each batch to contain an adequate amount of data from all the classes, which ensures each class receives appropriate supervision signals \cite{wang2017learning, kang2019decoupling, zhu2022balanced}. We consider resampling categories dynamically in each batch to construct contrastive pairs. For analyzing patents in the constrastive learning framework, we use a positive sample as the patent image and its generated captions, and the negative sample as the patent image with the caption of another patent. The quality of negative image-text pairs plays a vital role in contrastive learning. Intuitively, our goal is twofold: (i) to maintain a balanced representation of all classes during the batch training, and (ii) while constructing batches, we also aim to increase the likelihood that each batch contains a more diverse set of classes, with a particular focus on sufficient representation from the tail classes.  To do so, we define the probability of sampling a class $c$ as,
\begin{equation}
p_c = \frac{\frac{1}{f_c^\beta}}{\sum_{j=1}^{C} \frac{1}{f_j^\beta}}, \label{eq:sample}
\end{equation}
where is $f_c$ the frequency of class $c$ in the batch, $C$ is the total number of classes, and $\beta$ is a hyperparameter of rebalancing the weights of the tail classes.

 The advantage of our definition of $p_c$ is that when an anchor from a tail class is involved, we can easily adjust the sampling process by increasing the value of $\beta$ in Eq. \ref{eq:sample} for the head (popular) classes. This reduces the likelihood of the negative pairs mainly originating from the head classes, and thereby prevents the tail classes from being overly penalized as negatives. This adjustment helps the model to learn better representations for the tail classes and leads to improved performance in downstream tasks like patent classification and patent retrieval.

\subsubsection{Class-aware Contrastive Loss} In addition to Class-aware Sampling, we introduce a modified Class-aware Contrastive Loss that further mitigates the imbalance by incorporating class-dependent weighting into the CLIP contrastive loss in Eq. \ref{eq:clip}, as follows:

\begin{equation}
\mathcal{L}_{\text{CACL}} = - \frac{1}{f_i^\beta} \log \frac{\exp(\text{sim}(\mathbf{v}_i, \mathbf{t}_i)/\tau)}{\sum_{k=1}^{N} \exp(\text{sim}(\mathbf{v}_i, \mathbf{t}_k)/\tau)}, \label{eq:mclip}
\end{equation}

where, $f_i$ is the frequency of class which the $i$-th sample belongs. This class-aware weighting ensures that the loss function penalizes misclassifications more heavily for tail classes, which encourages the model to allocate more capacity to learn these classes.

\subsection{Multi-task Pre-training}

Considering the properties of patent data and tasks, we train \ourmodel on three tasks simultaneously, including \textit{(1)} class-aware image-generated description contrastive learning, \textit{(2)} class-aware classification, and \textit{(3)} class-aware image-image (other views) contrastive learning. In task (1), we use the methods in Class-aware Learning. In this section, we introduce task (2) and task (3).

\subsubsection{Class-aware Classification} In this task, we aim to handle the class imbalance in addition to the Class-aware Learning. It helps to capture the subtle differences between patents which enhances the model's ability to generalize across diverse categories. Specifically, \ourmodel incorporates an attention-based \cite{vaswani2017attention} classification layer with both image and text features. Let $v_i \in \mathbb{R}^{d_v}$ denote the visual features extracted from the \(i\)-th image by an image encoder, and $t_i \in \mathbb{R}^{d_t}$ denote the textual features extracted by a text encoder, the attention scores is computed as $
\alpha_i = \text{softmax}\left(W_v v_i + W_t \mathbf{t}_i + b_a\right)$, and then the combined multimodal vector $h_i$ is calculated by $
h_i = \alpha_i \odot \left(W_v' v_i + W_t' t_i\right)$, where $W$ are learnable weights and $b$ is bias. Finally, the class-aware classification loss is computed as:

\begin{equation}
\mathcal{L}_{\text{CACLS}} = - \frac{1}{f_i^\beta} \log \frac{\exp(w_c^\top h_i)}{\sum_{j=1}^{C} \exp(w_j^\top h_i)},\label{eq:cacls}
\end{equation}
where $f_i$ is the frequency of class in which $i$-th sample belongs, $\beta$ is a hyperparameter of rebalancing the weights of each class, and $w_c$ represents the weight vector of class $c$.

This class-aware classification approach addresses the class imbalance to ensure that \ourmodel effectively pays importance to the underrepresented (tail) classes. We also add the attention mechanism to learn the most relevant features in both image and text modalities. Moreover, by applying class-aware weighting, \ourmodel can be generalized across a wide variety of categories of design patents.


\subsubsection{Multi-view Image-image Contrastive Learning} Patents often include multiple views of the same design to provide a comprehensive understanding of the object from different angles, as shown in Figure \ref{fig:pipeline} (``Multi-view Images"). To effectively learn the multi-view information, we employ a similar class-aware contrastive learning approach in Class-aware Learning. The goal is to learn consistent representations across different views of the same patent while distinguishing these from views of other patents. Note that, the number of views varies in each patent, and some contain over 10 views. Thus, we mainly focus on front, top and side views. By aligning features from different views of the same object in a shared embedding space, the model gains a better understanding of the design. Similar to Eq. \ref{eq:mclip}, given image-image pairs $\{(v_i,v_i')\}^M_{i=1}$, the multi-view image-image contrastive learning loss is written as: 

\begin{equation}
\mathcal{L}_{\text{MVCL}} = - \frac{1}{f_i^\beta} \log \frac{\exp(\text{sim}(\mathbf{v}_i, \mathbf{v}_i')/\tau)}{\sum_{k=1}^{M} \exp(\text{sim}(\mathbf{v}_i, \mathbf{v}_k')/\tau)},\label{eq:mvcl}
\end{equation}
where $f_i$ is the frequency of class which the $i$-th sample belongs. To construct positive pairs, we randomly sample the image from the front view, side view, and top view.


Finally, our pre-train loss of \ourmodel is a linear combination of three tasks as follows: 

\begin{equation}
    \mathcal{L} = \lambda_1 \mathcal{L}_{\text{CACLS}} + \lambda_2 \mathcal{L}_{\text{CACL}} +\lambda_3 \mathcal{L}_{\text{MVCL}}
\end{equation}


 




\section{Experiments}
\textbf{Reproducibility.} We make the codebase available here: \url{https://anonymous.4open.science/r/PATENTCLIP-4B3F/README.md}
\
\subsection{Dataset}
We use a total of 285,391 patents for training \ourmodel, and 10,000 patents for validation from IMPACT dataset \cite{shomee2024impact}. Moreover, we also generate captions following IMPACT for different views with the prompt: \textit{This is the \{patent\_view\} view image of \{patent\_title\}. What is the shape of the image? What is the functionality of \{patent\_title\}?} The ``patent\_view'' includes top, front, and side views. In addition, we used data including 22,467 design patents from 2023 as test set, such as evaluations on zero-shot performance. Note that we use IMPACT as a baseline for patent downstream tasks. 

\subsection{Implementation details}
We use an open source implementation of CLIP\footnote{\url{https://github.com/mlfoundations/open_clip}}. The backbone models are ResNet50, ResNet101, ViT-B-32 and ViT-L-14. The hyperparameters for the best performance are listed as follows: learning rate is $5e-6$, weight decay is 0.1, optimizer is AdamW for all models, $\beta$ is 1.2, $\lambda_1$ is 1, $\lambda_2$ is 0.1 and $\lambda_3$ is 0.2. The batch size is 128, except 64 for ViT-L-14. We use image size of 224 $\times$ 224 in all the experiments. All pre-training experiments are conducted on a cluster of 4 NVIDIA A40 GPUs, and the downstream tasks and ablation studies are on 4 NVIDIA V100 GPUs.

\subsection{Patent downstream tasks}
In this work, we mainly demonstrate \ourmodel is beneficial for patent domain tasks, including patent classification and retrieval. However, there are limited studies focus on design patent, and they only work on image retrieval task. To evaluate the effectiveness of \ourmodel, we consider to conduct experiments on image retrieval comparing with the state-of-the-art methods, and evaluate on design patent classification and multimodal retrieval comparing with the general CLIP which is pre-trained on natural images and IMPACT results.

\subsubsection{Image retrieval}
Recent design patent studies mainly focus on image retrieval (IR). This process is often used for discovering new patents and assessing their novelty. These studies are validated with DeepPatent dataset \cite{Kucer_2022_WACV} containing 45,000 design patents. To verify \ourmodel, we conduct experiments compare to the SOTA model \cite{higuchi2023patent} on DeepPatent. We reproduce the same retrieval pipeline which used ArcFace \cite{deng2019arcface} with different backbones. In our settings, we utilize \ourmodel as backbones and also compare the performance of only using vision features and combining vision and textual features. More details of the implementations are shown in the Appendix (Sec. \ref{app:implementation_detail}). 

Table \ref{tab:ir} illustrates the results of image retrieval. The evaluation metric is the mean Average Precision (mAP) over all queries in the test set. \ourmodel outperforms SOTA models by replacing the backbones which are pre-trained on patent data with our proposed methods. Indeed, textual features are also beneficial to image retrieval. We show the retrieval examples in the Appendix.

\begin{table}[]
\caption{Image retrieval results (mAP) comparison between \ourmodel and other patent SOTA models on DeepPatent test set. Image size is 224 $\times$ 224. The best results are in bold. *Denotes our implementation.}
\label{tab:ir}
\centering
\resizebox{\columnwidth}{!}{\begin{tabular}{cc}
\toprule
\bf{Model}                & \bf{mAP} \\
\midrule
DeepPatent \cite{Kucer_2022_WACV}          &     0.379\\
ViT-B + ArcFace \cite{higuchi2023patent}      &     0.614\\
SWIN + ArcFace \cite{higuchi2023patent}      &     0.676\\
CLIP-ViT-B + ArcFace*  &     0.645\\
IMPACT \cite{shomee2024impact} & 0.657\\
\midrule
\textbf{\ourmodel-ViT-B (image)} + ArcFace            &  \bf{0.698}   \\
\textbf{\ourmodel-ViT-B (image+text)} + ArcFace            & \bf{0.712}   \\
\bottomrule
\end{tabular}
}

\end{table}

\subsubsection{Patent classification}
Patent classification is important but time-consuming for patent reviewers, but classifying design patents with AI method is unexplored. Thus, we consider to showcase \ourmodel on the classification task. The baseline model is OpenAI pre-trained CLIP and IMPACT. We use our test set (patents from 2023) to demonstrate the classification tasks, including zero-shot classification, and finetuning with a linear classifier. More details of the implementations are shown in the Appendix (Sec. \ref{app:implementation_detail}).

We demonstrate two settings of patent classification results in terms of accuracy in Table \ref{tab:classfication}. We use backbones with RN101 and ViT-B to compare the results. In all settings, \ourmodel performed better than CLIP and IMPACT. Notably, in fine-tuning settings, \ourmodel  outperforms CLIP by 60.1\% for RN101 and by 5.4\% for ViT-B, and achieves a 35.9\% improvement for RN101 and a 3.6\% improvement for ViT-B comparing IMPACT. Therefore, \ourmodel provides better representations for patent classification. 

\begin{table}[ht]
\caption{Accuracy (\%) for patent classification: comparison between CLIP and \ourmodel on 2023 test set. The best results are highlighted in bold in both settings.} \label{tab:classfication}
\resizebox{\columnwidth}{!}
{\begin{tabular}{cccc}
\toprule
\bf{Model}      & \bf{Backbone} & \bf{Zero-shot} & \bf{Fine-tune} \\
\midrule
\multirow{2}{*}{CLIP}       & RN101    & 11.91          & 22.45          \\
           & ViT-B    &  10.88          &     43.06       \\
\multirow{2}{*}{IMPACT*}       & RN101    & 11.89          & 27.66          \\
           & ViT-B    &  12.39          &     43.81       \\
\midrule
\multirow{2}{*}{\bf{\ourmodel}}  &  RN101        &      11.93     &       35.93    \\
           &   ViT-B       &  \bf{14.70}         &    \bf{45.37}      \\
\bottomrule
\end{tabular} }
\end{table}

\subsubsection{Multimodal retrieval}
The patent retrieval task is to identify relevant patent documents and images in response to search queries. In this task, we focus on multimodal retrieval, which incorporates both text and images. This integration enhances the ability to cross-reference and verify information, thus improving the overall effectiveness and efficiency of patent searches. In addition, multimodal retrieval can enable creativity and innovation in design by providing richer and more diverse sources of inspiration. We perform experiments on zero-shot text-image (T2I) and image-text (I2T) retrieval tasks on our validation set. 

We evaluate multimodal retrieval performance (Recall@K\footnote{The metric R@K evaluates whether the ground truth appears within the top K results of the validation set.}) and present results in Table \ref{tab:patentclipfulldata}. It shows that \ourmodel outperforms CLIP with all backbones on two tasks. As shown in the results, \ourmodel gains significant improvements, by up to \textit{360\%} and \textit{348\%} respectively at R@5 on T2I and I2T retrievals. Note that, ViT-L obtains the best recall in all settings, which demonstrates that the larger advanced models boost the performance. Multimodal retrieval examples are provided in the Appendix.


\begin{table}[]
\caption{Multimodal retrieval performance comparison between CLIP, IMPACT and \ourmodel on validation set. The ViT-L-14 model demonstrated superior performance over the other three backbone models tested. The highest Recall@K (\%) values are highlighted in bold. *Denotes our implementations.} \label{tab:patentclipfulldata}

\resizebox{\columnwidth}{!}{%
\begin{tabular}{cccccc}
\toprule
\multirow{2}{*}{\textbf{Model}} & \multirow{2}{*}{\textbf{Backbone}} & \multicolumn{2}{c}{\textbf{Text-Image}} & \multicolumn{2}{c}{\textbf{Image-Text}} \\
&   & R@5 & R@10  & R@5   & R@10               \\
\midrule
\multirow{4}{*}{CLIP}  & RN50  &   5.47&    8.51&      5.24&         7.72\\
 & RN101   & 7.60& 11.17& 6.10& 9.35\\
 & ViT-B   &   7.49&   10.60&   6.90&   10.34\\
& ViT-L    &   13.26&   18.29&   12.07&   17.17\\
\multirow{4}{*}{IMPACT*}  & RN50 & 17.21& 23.18 &14.67&    21.48\\
 & RN101   & 22.10& 31.35 & 20.32& 27.70\\
 & ViT-B   &   25.60&   34.92&   24.88&   35.12\\
& ViT-L    & 37.34&   50.56&   38.79&   51.05\\
\midrule
\multirow{4}{*}{\bf{\ourmodel}}  & RN50  & 25.17  & 34.50  & 23.49 & 32.70 \\
 & RN101 & 26.71 & 36.51   & 25.37 & 34.84  \\
& ViT-B   & 29.75   & 39.91   & 28.39   & 38.26   \\
& \bf{ViT-L}  & \bf{42.30}  & \bf{52.80}  & \bf{40.14}  & \bf{53.98}  \\
\bottomrule
\end{tabular}
}

\end{table}


\subsection{Ablation studies} 
We perform the ablation studies to analyze the hyperparameter settings and the components of \ourmodel, including the effectiveness of captions, multiple views, and pre-train tasks. Considering the limitations of computing resources, we use the patents from the recent five years for all ablation studies, including 113,887 patents in the train set and 5,000 patents in the validation set. We perform all ablation studies on the backbone of ViT-B-32. We also include more detailed ablation studies in the Appendix (Sec. \ref{app:ablation}).

\subsubsection{Analysis of detailed descriptions} 
We further analyze on the effectiveness of different texts. Specifically, we follow IMPACT dataset and classify their generated descriptions into captions only with ``Patent Title'', captions only with ``Shape'' and full captions. Based on the results shown in Table \ref{tab:text}, the detailed descriptions, including the shape and functions are beneficial for multimodal models to adjust to the patent domain. A notable increase of over 3$\%$ on both multimodal retrieval tasks indicates that \ourmodel, which is trained with generated descriptions has, a better understanding for further assessing and inspiring on design patents. 

\subsubsection{Impact of different views} We illustrate different views results of text-image retrieval and image retrieval, as shown in Table \ref{tab:view}. Front views benefits both retrieval tasks with $\approx$ 3 $\%$ and 1.8 $\%$ increase respectfully. However, our experiments demonstrate that not all views contribute equally to the performance of retrieval tasks. Using only side and top views can significantly reduce the model's generalization ability. We believe these views may not capture the most distinguishing features of the design, leading to confusion when the model attempts to match these views with corresponding text or images. Case studies are shown in the Appendix (Sec. \ref{sec:diff_views}).

\begin{table}[h]

\centering
\caption{Ablation studies on different text and views. Backbone models is ViT-B. The best results of multimodal retrieval (Recall@K (\%)) and image retrieval (mAP) are in bold.}
\begin{subtable}[h]{0.45\columnwidth}
\small
\centering
\begin{tabular}{ccc}

\toprule
\multirow{2}{*}{\bf{Text}} & \bf{I2T} & \bf{T2I} \\ 
                      & R@5 & R@5 \\ \midrule
Patent Title                 & 22.38   & 22.18    \\
Shape                 &  23.80    & 24.16    \\
Full captions         & \bf{25.56}    &   \bf{25.90}  \\ \bottomrule
\end{tabular}
\caption{ Text inputs }
\label{tab:text}
\end{subtable}
\hfill
\begin{subtable}[h]{0.45\columnwidth}
\small
\centering
\begin{tabular}{ccc}

\toprule
\multirow{2}{*}{\bf{Views}} & \bf{T2I} & \bf{IR}  \\ 
                       & R@5 & mAP \\ \midrule
Side                   &  8.24   & 0.611    \\
Top                    &  9.12   &  0.605    \\ 
Front                  &  \bf{12.05}   &  \bf{0.629}   \\ \bottomrule
\end{tabular}
\caption{ Views }
\label{tab:view}
\end{subtable}

\end{table}

\subsubsection{Effectiveness of pre-train tasks} To verify the effectiveness of our proposed multi-task pre-training, we conduct experiments on different combinations of pre-train tasks. Pre-training on all tasks brings significant improvements for all downstream tasks in zero-shot settings, as results shown in Table \ref{tab:pre-train}. Thus, \ourmodel not only enhances the model's generalization capabilities but also can be more adaptable to the patent tasks.

\begin{table}[]
\centering
\caption{Ablation studies on pre-train tasks of \ourmodel. We evaluate classification (Accuracy ($\%$)), multimodal retrieval (Recall@K (\%)) and image retrieval (mAP) with ViT-B. The best results are in bold.}
\resizebox{\columnwidth}{!}{%
\begin{tabular}{ccccccc}
\toprule
\multicolumn{3}{c}{\bf{Pre-train tasks}}                     &  \bf{Classification}&\bf{I2T}&  \bf{T2I}&\bf{IR}\\ \midrule
(1) & (2) & (3) &  Accuracy&R@5                  &  R@5                  &mAP             \\
\checkmark &   &             &                       3.18&20.94&                  21.24&0.649\\
\checkmark & \checkmark &             &                       8.29&21.38&                  20.98&0.652\\
\checkmark & \checkmark & \checkmark &                       \bf{8.64}&\bf{21.56}&                 \bf{21.71}&\bf{0.658}\\
\bottomrule
\end{tabular}} 

\label{tab:pre-train}
\end{table}

\subsubsection{Hyperparameters Analysis} 
\begin{figure}        
     \begin{subfigure}[b]{0.232\textwidth}
         \includegraphics[width=\textwidth]{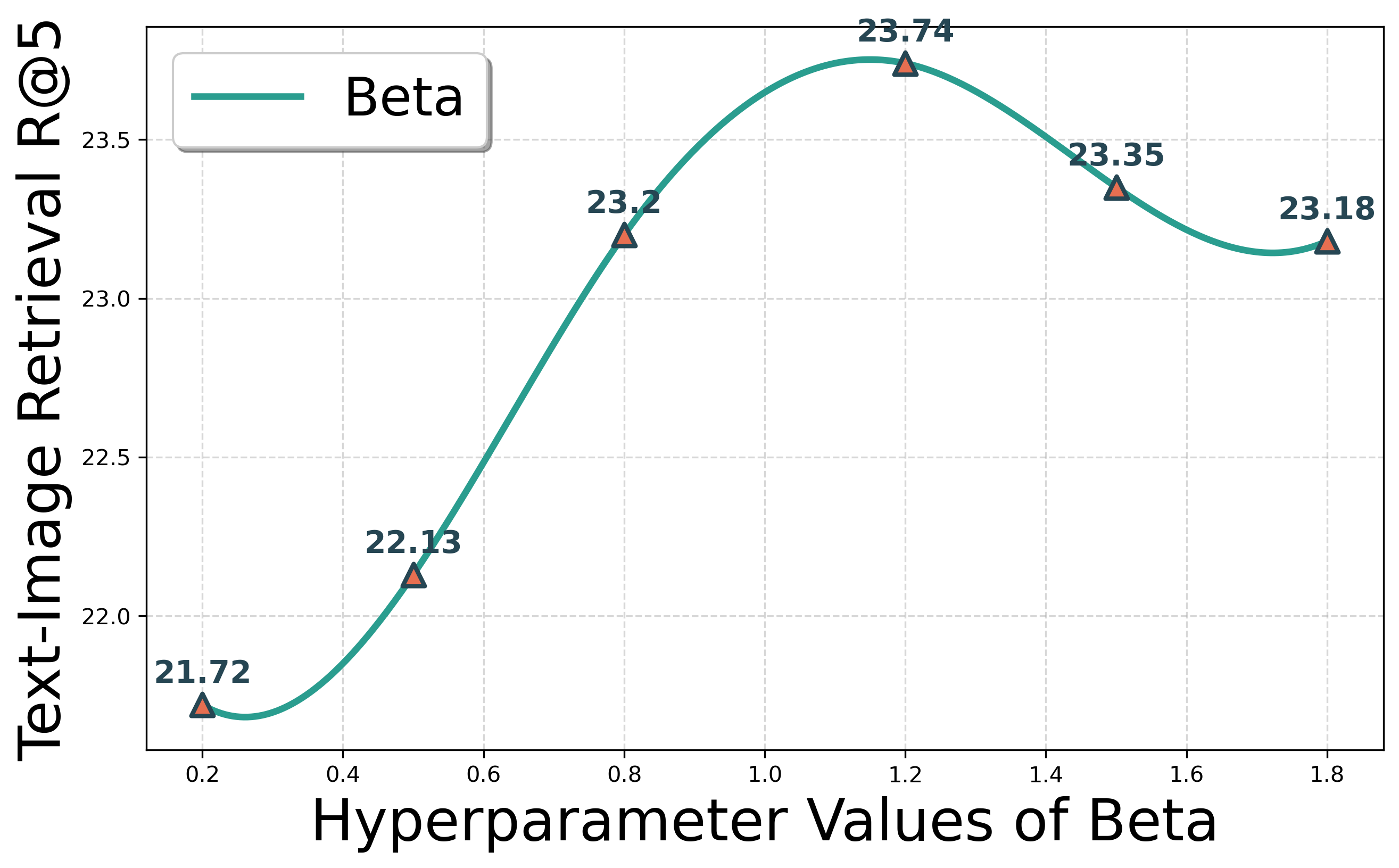}
         \caption{T-I R@5 with different $\beta$}
         \label{fig:beta}
    \end{subfigure}
    \begin{subfigure}[b]{0.232\textwidth}
         \includegraphics[width=\textwidth]{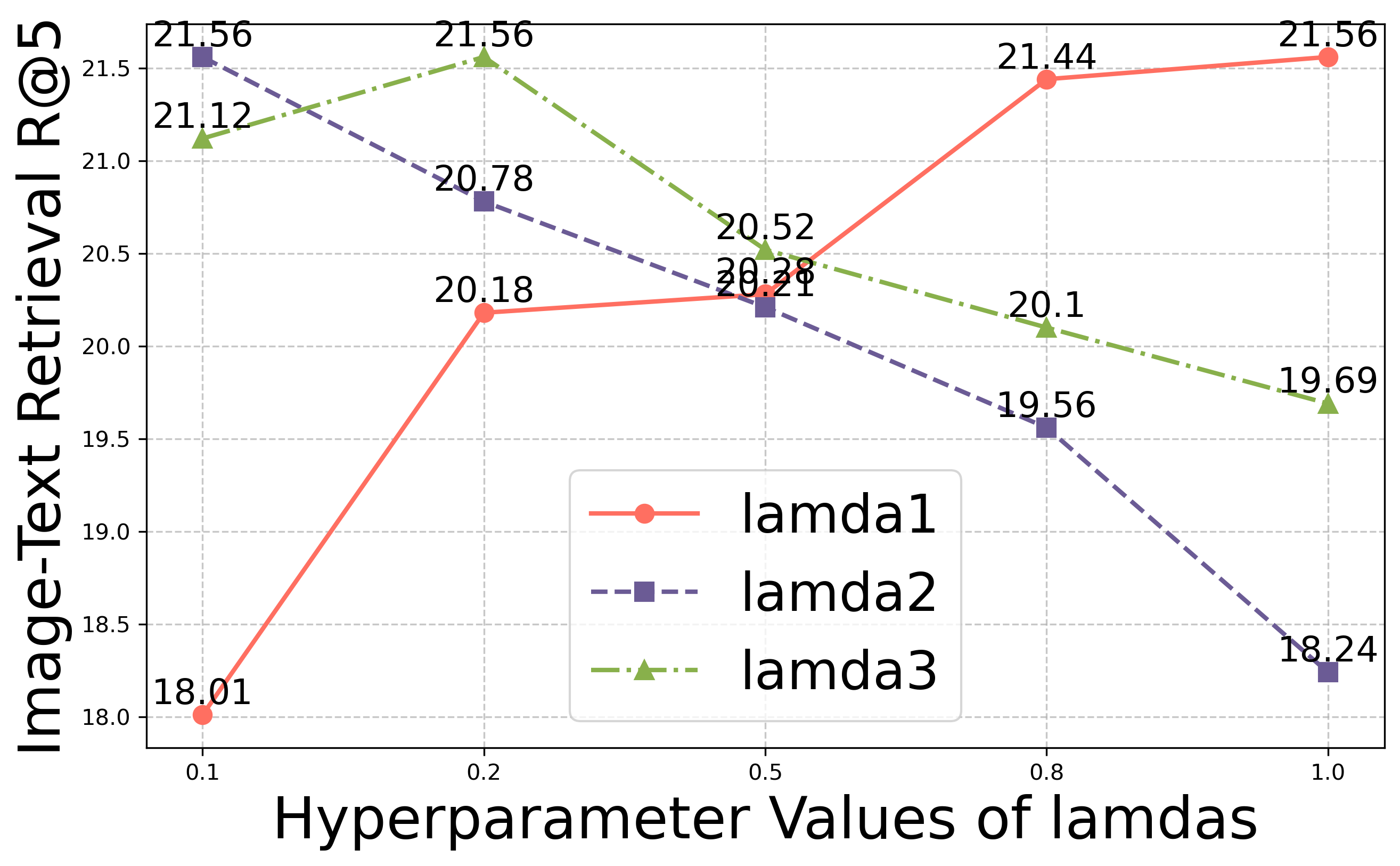}
         \caption{I-T R@5 with different $\lambda$s}
         \label{fig:lamda}
    \end{subfigure}

\caption{Ablation on hypermateters of $\beta$ and $\lambda$s.}
\vspace{-5mm}
\label{fig:hyper}
\end{figure}

As results shown in Figure \ref{fig:hyper}, we can conclude \textit{(1)} increasing $\beta$ values for balancing tail classes can improve the performance, see Figure \ref{fig:beta}, but bigger $\beta$ values may cause overfit on the tail classes which will be harmful for the overall pre-training. \textit{(2)} Figure \ref{fig:lamda} shows that the weight assgined to $\mathcal{L}_{\text{CACL}}$ is crucial to the performance. $\mathcal{L}_{\text{CACLS}}$ ensures that the model learns to categorize images correctly, but emphasis on the classification could reduce the model's ability to generalize to new data. Similar to the results in impact of different views, while the multi-views help in learning consistent representations across different views, a smaller weight for $\mathcal{L}_{\text{MVCL}}$ indicates that image views alignment needs to be carefully considered and may diminish the benefits.


\begin{figure}[ht]
        
     \begin{subfigure}[b]{0.23\textwidth}
         \includegraphics[width=\textwidth]{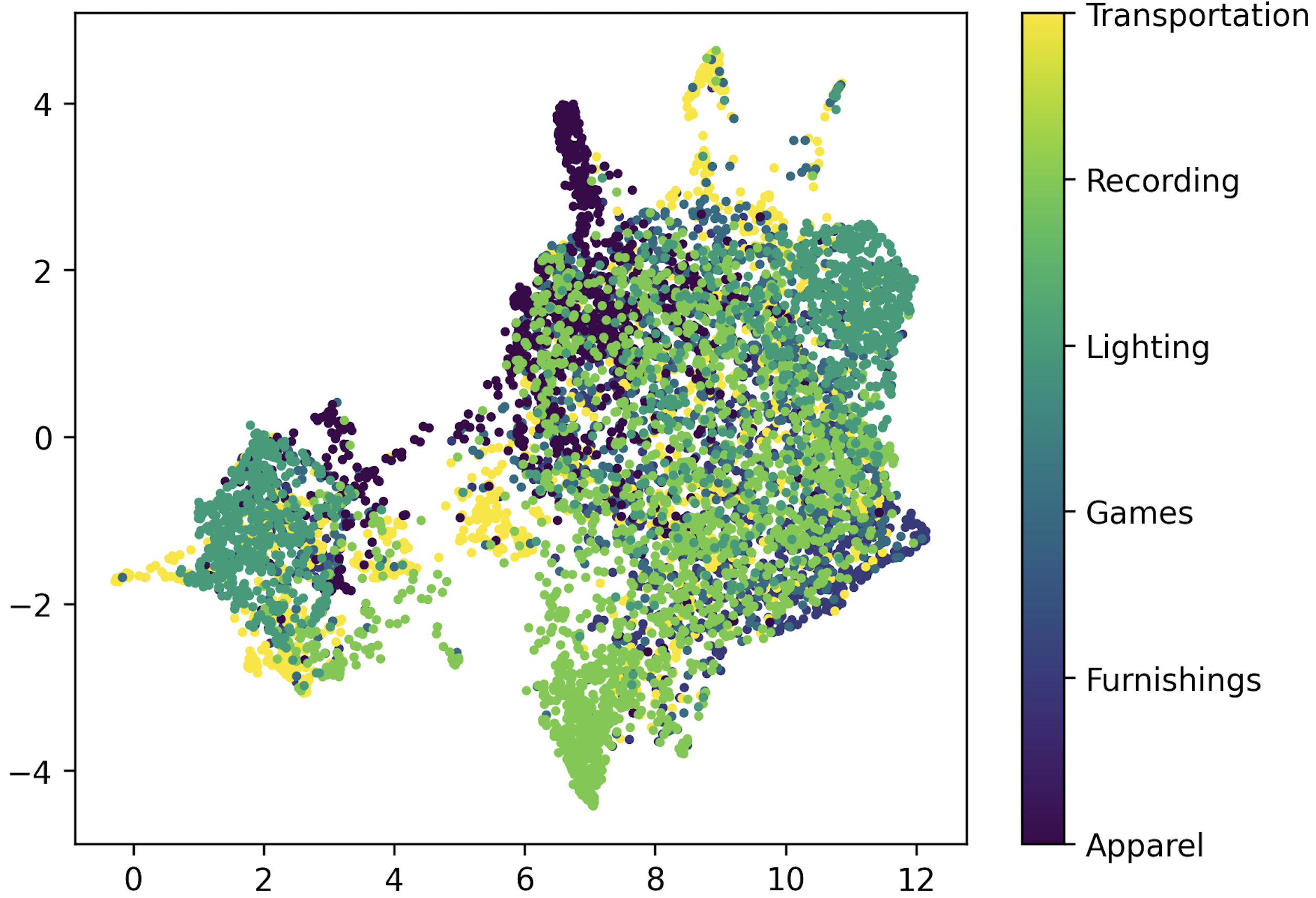}
         \caption{CLIP}
         \label{fig:clipi}
    \end{subfigure}
    \hfill
    \begin{subfigure}[b]{0.23\textwidth}
         \includegraphics[width=\textwidth]{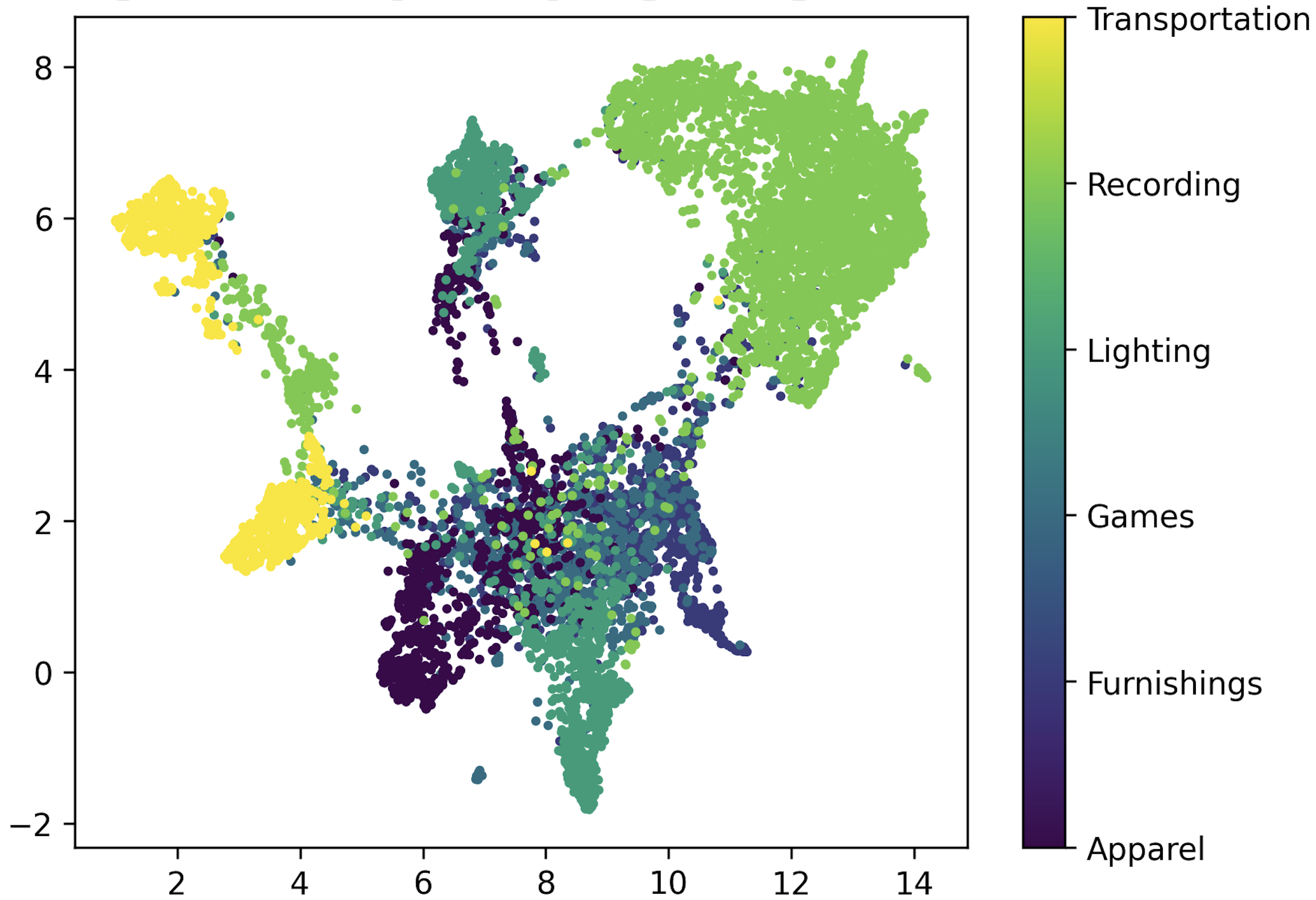}
         \caption{\ourmodel}
         \label{fig:clipt}
    \end{subfigure}

\caption{UMAP feature embeddings for patent images. (a) Visualization of features using CLIP models  (b) Visualization of features using \ourmodel.  \ourmodel shows well formed clusters in image features.} \label{fig:feature}
\vspace{-5mm}
\end{figure}

\subsection{Qualitative Analysis}

\textbf{U-MAP projection analysis. }Figure \ref{fig:feature} shows the learned image features for sample patents using U-MAP projection \cite{mcinnes2018umap-software}. Different colors represent the clusters of the corresponding classes. We observe that \ourmodel can identify clusters over the extracted image features, but CLIP is not able to classify the patent images. Therefore, we believe that \ourmodel is beneficial in the specific patent domain for many downstream tasks, such as classification and retrieval. 



\begin{figure}[ht]
         \centering
         \includegraphics[width=.99\columnwidth]{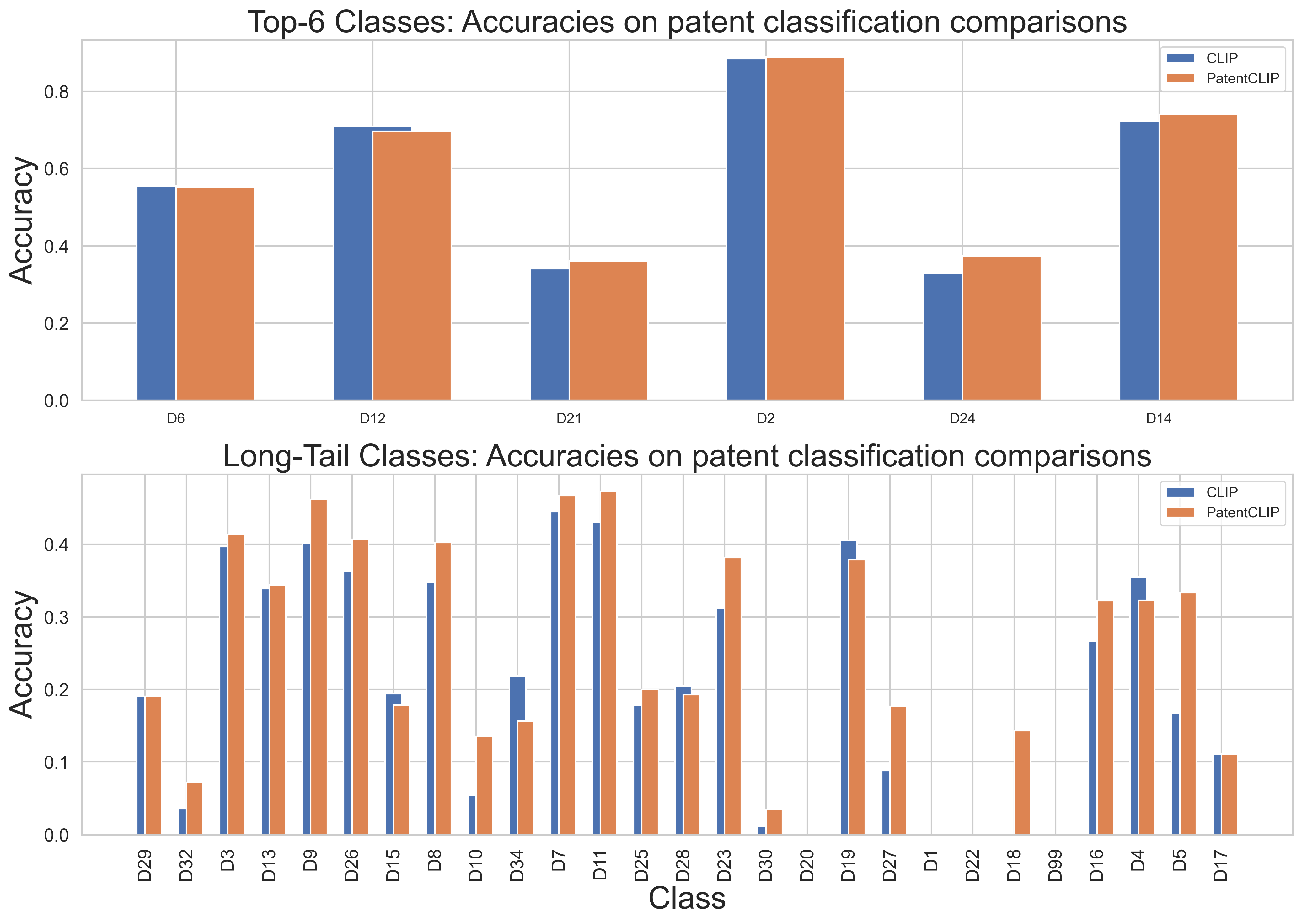}

        \caption{\ourmodel improve classification accuracies for 20 classes among long-tail distributions. }
        \vspace{-3mm}
        \label{fig:class_comp}
\end{figure}

\textbf{Class-aware classification analysis:} We further analyze the impact of class-aware learning in \ourmodel. On patent classification tasks, \ourmodel improves by 1.22 \% on the top-6 classes and 2.90 \% on the long-tail classes. Among all 33 classes, \ourmodel outperforms CLIP in 20 classes under fine-tuning (see Figure \ref{fig:class_comp}). These results highlight the importance and effectiveness of class-aware learning for improving performance on long-tail classes.

\begin{figure}[ht]
         \centering
         \includegraphics[width=.99\columnwidth]{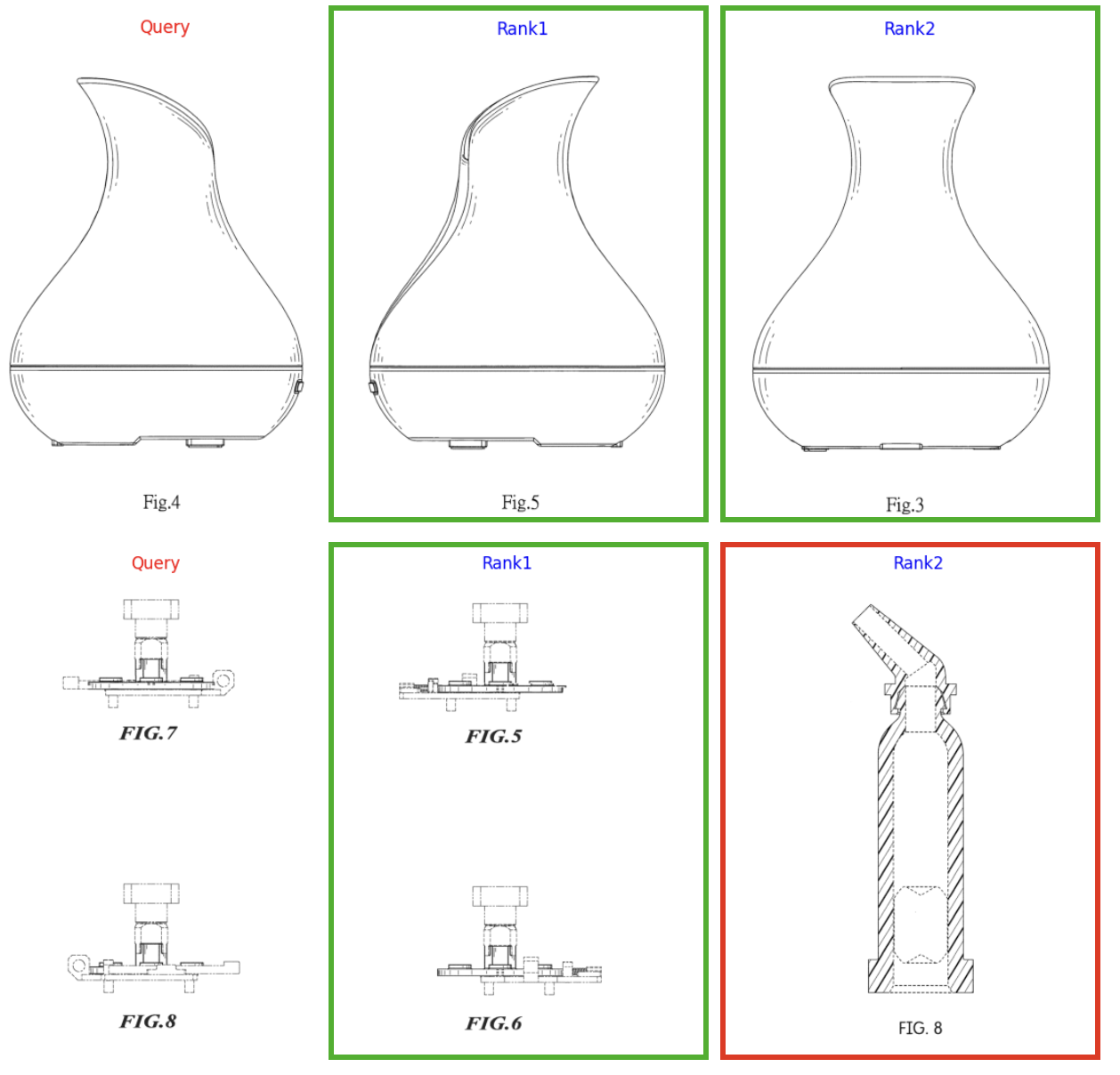}

        \caption{\ourmodel often retrieves different views as the top results (the first row). However, the results are not always accurate (red box in second row). }
        \vspace{-5mm}
        \label{fig:example_ir}
\end{figure}

We also show the retrieval results of multiple views from \ourmodel. Figure \ref{fig:example_ir} presents the top two retrievals. \ourmodel often retrieves the different views of the query figure. Figure \ref{fig:ir1} (see the Appendix) shows that CLIP often cannot retrieve multiple views of the same query figure.

\section{Conclusions} 

In this paper, we propose \ourmodel to provide a domain-aware multimodal model for design patents. We first consider class-aware sampling and contrastive learning for the long-tail distribution in design patent data. 
In addition, multi-view image-image contrastive learning provides a comprehensive understanding of the patents from different angles, which can improve the performance in patent retrieval. Finally, we pre-train CLIP-based models on multi-tasks tailored for design patents. Our proposed \ourmodel outperforms the baseline models on patent classification, image retrieval and multimodal retrievals. These demonstrate that \ourmodel can provide a better understanding and generalization ability for design patents in multimodal scenarios. 

\clearpage
\section{Ethical considerations}
The ethical considerations for the multimodal analysis with \ourmodel include the followings: 

\begin{itemize}
\item \textbf{Needs for human supervision. }
As with many AI tools, \ourmodel could be misused for non-scientific or adversarial purposes, such as generating infringing designs. The model is intended solely for research, analysis, and balancing the use of technology with human oversight is important to maintain the quality and integrity of patent applications. 

\item \textbf{Legal issues.} Ethical considerations should also include to ensure that the design patents generated from \ourmodel comply with legal requirements and regulations of patent laws.
\end{itemize}

\section{Limitations}
Patents often include a varying number of views, and many views are difficult to understand even for humans. Additionally, in many cases, patents have new designs on the front view but contain similar side and top views, which challenge the model to distinguish the patterns between different patents. This can lead to a poor alignment and a less effective learning. Thus, we will consider to address the challenges of incorporating multi-views as a future research direction.
\clearpage

\bibliography{main}

\begin{thebibliography}{38}
\providecommand{\natexlab}[1]{#1}

\bibitem[{Achiam et~al.(2023)Achiam, Adler, Agarwal, Ahmad, Akkaya, Aleman, Almeida, Altenschmidt, Altman, Anadkat et~al.}]{achiam2023gpt}
Josh Achiam, Steven Adler, Sandhini Agarwal, Lama Ahmad, Ilge Akkaya, Florencia~Leoni Aleman, Diogo Almeida, Janko Altenschmidt, Sam Altman, Shyamal Anadkat, and 1 others. 2023.
\newblock Gpt-4 technical report.
\newblock \emph{arXiv preprint arXiv:2303.08774}.

\bibitem[{Deng et~al.(2009)Deng, Dong, Socher, Li, Li, and Fei-Fei}]{deng2009imagenet}
Jia Deng, Wei Dong, Richard Socher, Li-Jia Li, Kai Li, and Li~Fei-Fei. 2009.
\newblock Imagenet: A large-scale hierarchical image database.
\newblock In \emph{2009 IEEE conference on computer vision and pattern recognition}, pages 248--255. Ieee.

\bibitem[{Deng et~al.(2019)Deng, Guo, Xue, and Zafeiriou}]{deng2019arcface}
Jiankang Deng, Jia Guo, Niannan Xue, and Stefanos Zafeiriou. 2019.
\newblock Arcface: Additive angular margin loss for deep face recognition.
\newblock In \emph{Proceedings of the IEEE/CVF conference on computer vision and pattern recognition}, pages 4690--4699.

\bibitem[{Dosovitskiy et~al.(2020)Dosovitskiy, Beyer, Kolesnikov, Weissenborn, Zhai, Unterthiner, Dehghani, Minderer, Heigold, Gelly et~al.}]{dosovitskiy2020image}
Alexey Dosovitskiy, Lucas Beyer, Alexander Kolesnikov, Dirk Weissenborn, Xiaohua Zhai, Thomas Unterthiner, Mostafa Dehghani, Matthias Minderer, Georg Heigold, Sylvain Gelly, and 1 others. 2020.
\newblock An image is worth 16x16 words: Transformers for image recognition at scale.
\newblock \emph{arXiv preprint arXiv:2010.11929}.

\bibitem[{Fall et~al.(2003)Fall, T{\"o}rcsv{\'a}ri, Benzineb, and Karetka}]{fall2003automated}
Caspar~J Fall, Atilla T{\"o}rcsv{\'a}ri, Karim Benzineb, and Gabor Karetka. 2003.
\newblock Automated categorization in the international patent classification.
\newblock In \emph{Acm Sigir Forum}, volume~37, pages 10--25. ACM New York, NY, USA.

\bibitem[{Ghauri et~al.(2023)Ghauri, M{\"u}ller-Budack, and Ewerth}]{ghauri}
Junaid~Ahmed Ghauri, Eric M{\"u}ller-Budack, and Ralph Ewerth. 2023.
\newblock Classification of visualization types and perspectives in patents.
\newblock In \emph{TPDL}.

\bibitem[{Gong et~al.(2024)Gong, Wang, Haurum, Lowe, Taylor, and Chang}]{bio}
ZeMing Gong, Austin~T Wang, Joakim~Bruslund Haurum, Scott~C Lowe, Graham~W Taylor, and Angel~X Chang. 2024.
\newblock Bioscan-clip: Bridging vision and genomics for biodiversity monitoring at scale.
\newblock \emph{arXiv preprint arXiv:2405.17537}.

\bibitem[{He et~al.(2016)He, Zhang, Ren, and Sun}]{he2016deep}
Kaiming He, Xiangyu Zhang, Shaoqing Ren, and Jian Sun. 2016.
\newblock Deep residual learning for image recognition.
\newblock In \emph{Proceedings of the IEEE conference on computer vision and pattern recognition}, pages 770--778.

\bibitem[{Hendriksen et~al.(2022)Hendriksen, Bleeker, Vakulenko, Van~Noord, Kuiper, and De~Rijke}]{ecommerce}
Mariya Hendriksen, Maurits Bleeker, Svitlana Vakulenko, Nanne Van~Noord, Ernst Kuiper, and Maarten De~Rijke. 2022.
\newblock Extending clip for category-to-image retrieval in e-commerce.
\newblock In \emph{European Conference on Information Retrieval}, pages 289--303. Springer.

\bibitem[{Higuchi et~al.(2023)Higuchi, Honbu, and Yanai}]{higuchi2023patent2}
Kotaro Higuchi, Yuma Honbu, and Keiji Yanai. 2023.
\newblock Patent image retrieval using cross-entropy-based metric learning.
\newblock In \emph{IW-FCV}.

\bibitem[{Higuchi and Yanai(2023)}]{higuchi2023patent}
Kotaro Higuchi and Keiji Yanai. 2023.
\newblock Patent image retrieval using transformer-based deep metric learning.
\newblock \emph{WPI}, 74:102217.

\bibitem[{Kamateri et~al.(2024)Kamateri, Salampasis, and Perez-Molina}]{kamateri2024will}
Eleni Kamateri, Michail Salampasis, and Eduardo Perez-Molina. 2024.
\newblock Will ai solve the patent classification problem?
\newblock \emph{World Patent Information}, 78:102294.

\bibitem[{Kamateri et~al.(2022)Kamateri, Stamatis, Diamantaras, and Salampasis}]{kamateri2022automated}
Eleni Kamateri, Vasileios Stamatis, Konstantinos Diamantaras, and Michail Salampasis. 2022.
\newblock Automated single-label patent classification using ensemble classifiers.
\newblock In \emph{ICMLC}.

\bibitem[{Kang et~al.(2019)Kang, Xie, Rohrbach, Yan, Gordo, Feng, and Kalantidis}]{kang2019decoupling}
Bingyi Kang, Saining Xie, Marcus Rohrbach, Zhicheng Yan, Albert Gordo, Jiashi Feng, and Yannis Kalantidis. 2019.
\newblock Decoupling representation and classifier for long-tailed recognition.
\newblock \emph{arXiv preprint arXiv:1910.09217}.

\bibitem[{Kang et~al.(2020)Kang, Lee, Lee, and Lee}]{kang2020patent}
Dylan~Myungchul Kang, Charles~Cheolgi Lee, Suan Lee, and Wookey Lee. 2020.
\newblock Patent prior art search using deep learning language model.
\newblock In \emph{IDEAS}.

\bibitem[{Krizhevsky et~al.(2012)Krizhevsky, Sutskever, and Hinton}]{krizhevsky2012imagenet}
Alex Krizhevsky, Ilya Sutskever, and Geoffrey~E Hinton. 2012.
\newblock Imagenet classification with deep convolutional neural networks.
\newblock \emph{Advances in neural information processing systems}, 25.

\bibitem[{Kucer et~al.(2022)Kucer, Oyen, Castorena, and Wu}]{Kucer_2022_WACV}
Michal Kucer, Diane Oyen, Juan Castorena, and Jian Wu. 2022.
\newblock Deeppatent: Large scale patent drawing recognition and retrieval.
\newblock In \emph{WACV}.

\bibitem[{Lei et~al.(2023)Lei, Li, Shen, Zhang, and Shan}]{med3}
Yiming Lei, Zilong Li, Yan Shen, Junping Zhang, and Hongming Shan. 2023.
\newblock Clip-lung: Textual knowledge-guided lung nodule malignancy prediction.
\newblock In \emph{International Conference on Medical Image Computing and Computer-Assisted Intervention}, pages 403--412. Springer.

\bibitem[{Li et~al.(2023)Li, Li, Savarese, and Hoi}]{blip2}
Junnan Li, Dongxu Li, Silvio Savarese, and Steven Hoi. 2023.
\newblock Blip-2: Bootstrapping language-image pre-training with frozen image encoders and large language models.
\newblock In \emph{International conference on machine learning}, pages 19730--19742. PMLR.

\bibitem[{Lin et~al.(2017)Lin, Goyal, Girshick, He, and Doll{\'a}r}]{lin2017focal}
Tsung-Yi Lin, Priya Goyal, Ross Girshick, Kaiming He, and Piotr Doll{\'a}r. 2017.
\newblock Focal loss for dense object detection.
\newblock In \emph{Proceedings of the IEEE international conference on computer vision}, pages 2980--2988.

\bibitem[{Liu et~al.(2021)Liu, Lin, Cao, Hu, Wei, Zhang, Lin, and Guo}]{liu2021swin}
Ze~Liu, Yutong Lin, Yue Cao, Han Hu, Yixuan Wei, Zheng Zhang, Stephen Lin, and Baining Guo. 2021.
\newblock Swin transformer: Hierarchical vision transformer using shifted windows.
\newblock In \emph{Proceedings of the IEEE/CVF international conference on computer vision}, pages 10012--10022.

\bibitem[{Lo et~al.(2024)Lo, Chu, Hsiang, and Cho}]{lo2024large}
Hao-Cheng Lo, Jung-Mei Chu, Jieh Hsiang, and Chun-Chieh Cho. 2024.
\newblock Large language model informed patent image retrieval.
\newblock \emph{arXiv preprint arXiv:2404.19360}.

\bibitem[{Lu et~al.(2024)Lu, Chen, Williamson, Chen, Liang, Ding, Jaume, Odintsov, Le, Gerber et~al.}]{med1}
Ming~Y Lu, Bowen Chen, Drew~FK Williamson, Richard~J Chen, Ivy Liang, Tong Ding, Guillaume Jaume, Igor Odintsov, Long~Phi Le, Georg Gerber, and 1 others. 2024.
\newblock A visual-language foundation model for computational pathology.
\newblock \emph{Nature Medicine}, 30(3):863--874.

\bibitem[{McInnes et~al.(2018)McInnes, Healy, Saul, and Grossberger}]{mcinnes2018umap-software}
Leland McInnes, John Healy, Nathaniel Saul, and Lukas Grossberger. 2018.
\newblock Umap: Uniform manifold approximation and projection.
\newblock \emph{The Journal of Open Source Software}, 3(29):861.

\bibitem[{Moser(2013)}]{moser2013patents}
Petra Moser. 2013.
\newblock Patents and innovation: evidence from economic history.
\newblock \emph{Journal of economic perspectives}, 27(1):23--44.

\bibitem[{M{\"u}ller et~al.(2022)M{\"u}ller, Kaissis, and Rueckert}]{med2}
Philip M{\"u}ller, Georgios Kaissis, and Daniel Rueckert. 2022.
\newblock The role of local alignment and uniformity in image-text contrastive learning on medical images.
\newblock \emph{arXiv preprint arXiv:2211.07254}.

\bibitem[{Rademaker(2000)}]{rademaker2000classification}
Charles~A Rademaker. 2000.
\newblock The classification of ornamental designs in the united states patent classification system.
\newblock \emph{World Patent Information}, 22(3):123--133.

\bibitem[{Radford et~al.(2021)Radford, Kim, Hallacy, Ramesh, Goh, Agarwal, Sastry, Askell, Mishkin, Clark et~al.}]{clip}
Alec Radford, Jong~Wook Kim, Chris Hallacy, Aditya Ramesh, Gabriel Goh, Sandhini Agarwal, Girish Sastry, Amanda Askell, Pamela Mishkin, Jack Clark, and 1 others. 2021.
\newblock Learning transferable visual models from natural language supervision.
\newblock In \emph{International conference on machine learning}, pages 8748--8763. PMLR.

\bibitem[{Rana et~al.(2023)Rana, Melnik, and S{\"u}nderhauf}]{robotics2}
Krishan Rana, Andrew Melnik, and Niko S{\"u}nderhauf. 2023.
\newblock Contrastive language, action, and state pre-training for robot learning.
\newblock \emph{arXiv preprint arXiv:2304.10782}.

\bibitem[{Sangkloy et~al.(2016)Sangkloy, Burnell, Ham, and Hays}]{sangkloy2016sketchy}
Patsorn Sangkloy, Nathan Burnell, Cusuh Ham, and James Hays. 2016.
\newblock The sketchy database: learning to retrieve badly drawn bunnies.
\newblock \emph{ACM Transactions on Graphics (TOG)}, 35(4):1--12.

\bibitem[{Shibata et~al.(2024)Shibata, Deguchi, and Taguchi}]{robotics1}
Kazuki Shibata, Hideki Deguchi, and Shun Taguchi. 2024.
\newblock Clip feature-based randomized control using images and text for multiple tasks and robots.
\newblock \emph{Advanced Robotics}, pages 1--13.

\bibitem[{Shomee et~al.(2024)Shomee, Wang, Ravi, and Medya}]{shomee2024impact}
Homaira~Huda Shomee, Zhu Wang, Sathya Ravi, and Sourav Medya. 2024.
\newblock Impact: a large-scale integrated multimodal patent analysis and creation dataset for design patents.
\newblock \emph{Advances in Neural Information Processing Systems}, 37:125520--125546.

\bibitem[{Siddharth et~al.(2022)Siddharth, Li, and Luo}]{siddharth2022enhancing}
L~Siddharth, Guangtong Li, and Jianxi Luo. 2022.
\newblock Enhancing patent retrieval using text and knowledge graph embeddings: a technical note.
\newblock \emph{Journal of Engineering Design}, 33:670--683.

\bibitem[{Sontakke et~al.(2024)Sontakke, Zhang, Arnold, Pertsch, B{\i}y{\i}k, Sadigh, Finn, and Itti}]{robotics3}
Sumedh Sontakke, Jesse Zhang, S{\'e}b Arnold, Karl Pertsch, Erdem B{\i}y{\i}k, Dorsa Sadigh, Chelsea Finn, and Laurent Itti. 2024.
\newblock Roboclip: One demonstration is enough to learn robot policies.
\newblock \emph{Advances in Neural Information Processing Systems}, 36.

\bibitem[{Sun et~al.(2024)Sun, Fan, Li, Wang, Ge, and Shang}]{edu}
Xiaoning Sun, Tao Fan, Hongxu Li, Guozhong Wang, Peien Ge, and Xiwu Shang. 2024.
\newblock Clip2tf: Multimodal video-text retrieval for adolescent education.
\newblock \emph{Displays}, page 102801.

\bibitem[{Vaswani et~al.(2017)Vaswani, Shazeer, Parmar, Uszkoreit, Jones, Gomez, Kaiser, and Polosukhin}]{vaswani2017attention}
Ashish Vaswani, Noam Shazeer, Niki Parmar, Jakob Uszkoreit, Llion Jones, Aidan~N Gomez, {\L}ukasz Kaiser, and Illia Polosukhin. 2017.
\newblock Attention is all you need.
\newblock In \emph{Advances in neural information processing systems}, pages 5998--6008.

\bibitem[{Wang et~al.(2017)Wang, Ramanan, and Hebert}]{wang2017learning}
Yu-Xiong Wang, Deva Ramanan, and Martial Hebert. 2017.
\newblock Learning to model the tail.
\newblock \emph{Advances in neural information processing systems}, 30.

\bibitem[{Zhu et~al.(2022)Zhu, Wang, Chen, Chen, and Jiang}]{zhu2022balanced}
Jianggang Zhu, Zheng Wang, Jingjing Chen, Yi-Ping~Phoebe Chen, and Yu-Gang Jiang. 2022.
\newblock Balanced contrastive learning for long-tailed visual recognition.
\newblock In \emph{Proceedings of the IEEE/CVF Conference on Computer Vision and Pattern Recognition}, pages 6908--6917.

\end{thebibliography}

\clearpage
\appendix


\clearpage
\setcounter{page}{1}
\section*{Appendix}

\section{Additional Implementation Details}
\label{app:implementation_detail}

We describe the implementation details of pre-training \ourmodel in our paper. In addition, we adjust specific requirements for patent downstream tasks, such as image retrieval and classification, and we will provide these implementation details in this section.

\subsection{Image retrieval}We integrate \ourmodel into SOTA patent image retrieval method \cite{higuchi2023patent}. We use pre-trained \ourmodel backbones and leave other settings the same. Specifically, the dataset used in this task is DeepPatent with patents from 2018 to 2019, including a train set of 254,787 images, a validation set of 44,815 images, and a test set of 38,834 images. The loss function is Focal loss \cite{lin2017focal}. The best hyperparameters are from \cite{higuchi2023patent} which are listed as follows: the batch size is 256, the optimizer is AdamW, the learning rate is $1e^{-4}$ and the number of training epochs is 25.

\subsection{Patent classification}
The design patents in our data has 33 classes. For the classification task, we use data from 2023. We show design categories and number of examples of our train data in Table \ref{tab:num_categories}. 
In patent classification, we finetune a simple linear classifier with frozen CLIP \cite{clip} and \ourmodel backbones (RN101 \cite{he2016deep} and ViT-B \cite{dosovitskiy2020image}) for comparisons. The hyerparameters are listed as follows: the batch size is 32, the optimizer is AdamW, the learning rate is $1e^{-4}$, and the number of training epochs is 15. 

\begin{table*}[!]
\centering
\caption{The table shows the list of U.S. design patent classes and number of occurrences in our classification task.}
\label{tab:num_categories}
\resizebox{\textwidth}{!}{
\begin{tabular}{clc}
\toprule
\textbf{Class} & \textbf{Description}& \textbf{Occurences} \\
\midrule
D1  & Edible Products& 38\\
D2  & Apparel and Haberdashery & 930\\
D3  & Travel Goods, Personal Belongings, and Storage or Carrying Articles & 462 \\
D4  & Brushware & 122\\
D5  & Textile or Paper Yard Goods; Sheet Material & \colorbox{brickred}{\textcolor{black}{20}} \\
D6  & Furnishings & 1052\\
D7  & Equipment for Preparing or Serving Food or Drink Not Elsewhere Specified & 906\\
D8  & Tools and Hardware & 735\\
D9  & Packages and Containers for Goods & 525 \\
D10 & Measuring, Testing or Signaling Instruments & 444 \\
D11 & Jewelry, Symbolic Insignia, and Ornaments &  369\\
D12 & Transportation & 1261\\
D13 & Equipment for Production, Distribution, or Transformation of Energy & 755\\
D14 & Recording, Communication, or Information Retrieval Equipment & \colorbox{green}{\textcolor{black}{1943}}\\
D15 & Machines Not Elsewhere Specified&  512\\
D16 & Photography and Optical Equipment& 359 \\
D17 & Musical Instruments& 35\\
D18 & Printing and Office Machinery& 55 \\
D19 & Office Supplies; Artists' and Teachers' Materials & 146\\
D20 & Sales and Advertising Equipment & 39\\
D21 & Games, Toys and Sports Goods& 962\\
D22 & Arms, Pyrotechnics, Hunting and Fishing Equipment & 184\\
D23 & Environmental Heating and Cooling, Fluid Handling and Sanitary Equipment& 743 \\
D24 & Medical and Laboratory Equipment& 1044 \\
D25 & Building Units and Construction Elements & 179 \\
D26 & Lighting& 901\\
D27 & Tobacco and Smokers' Supplies& 136\\
D28 & Cosmetic Products and Toilet Articles& 329 \\
D29 & Equipment for Safety, Protection and Rescue & 80\\
D30 & Animal Husbandry&  347\\
D32 & Washing, Cleaning or Drying Machines &221\\
D34 & Material or Article Handling Equipment & 127\\
D99 & Miscellaneous & 26\\
\bottomrule
\end{tabular}
}
\end{table*}

\begin{figure*}
     \centering

     \begin{subfigure}[b]{0.48\textwidth}
         \centering
         \includegraphics[width=\textwidth]{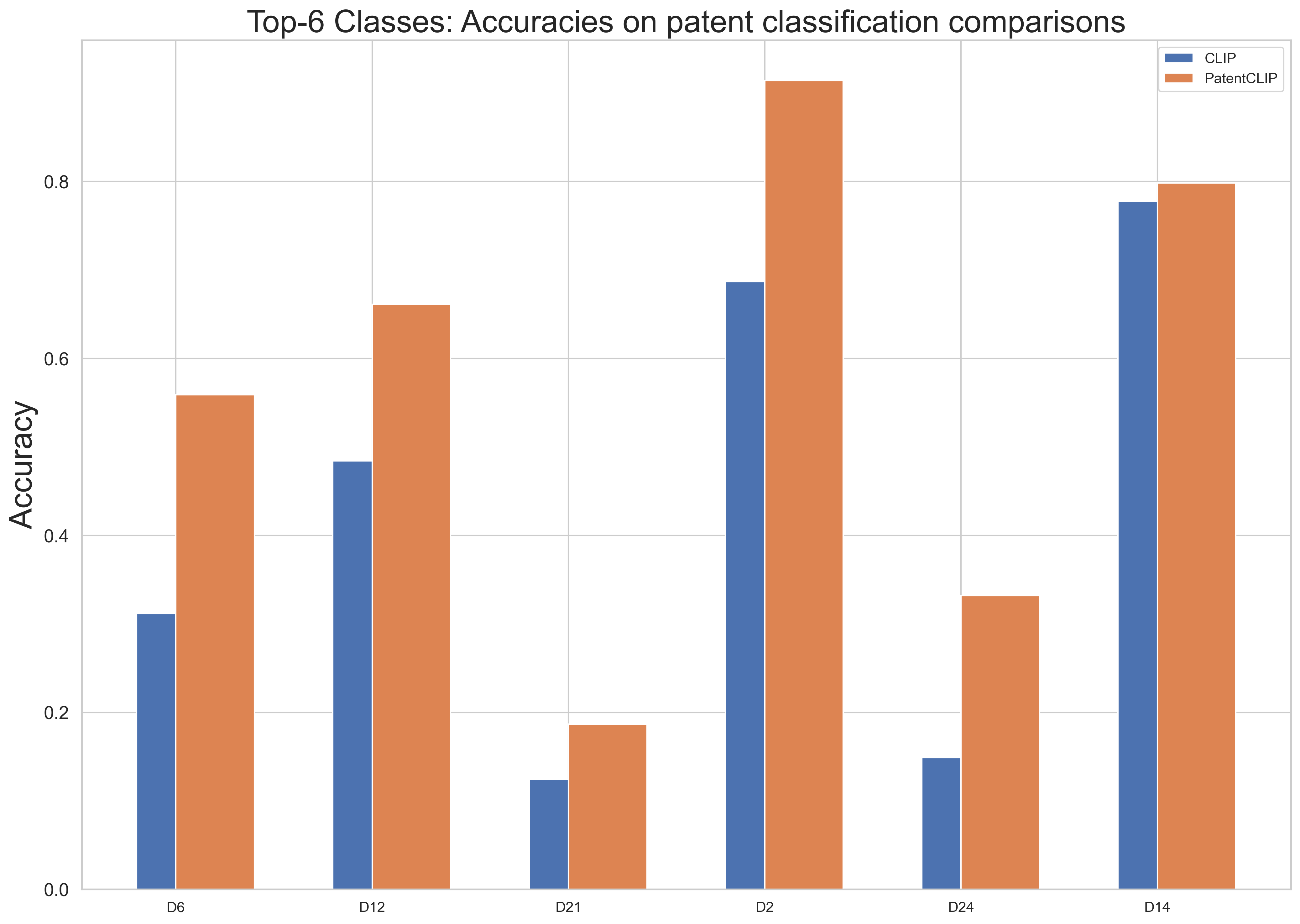}
         \caption{Patent classification comparisons on top 6 classes}
         \label{fig:rn101top}
     \end{subfigure}
     \hfill
     \begin{subfigure}[b]{0.48\textwidth}
         \centering
         \includegraphics[width=\textwidth]{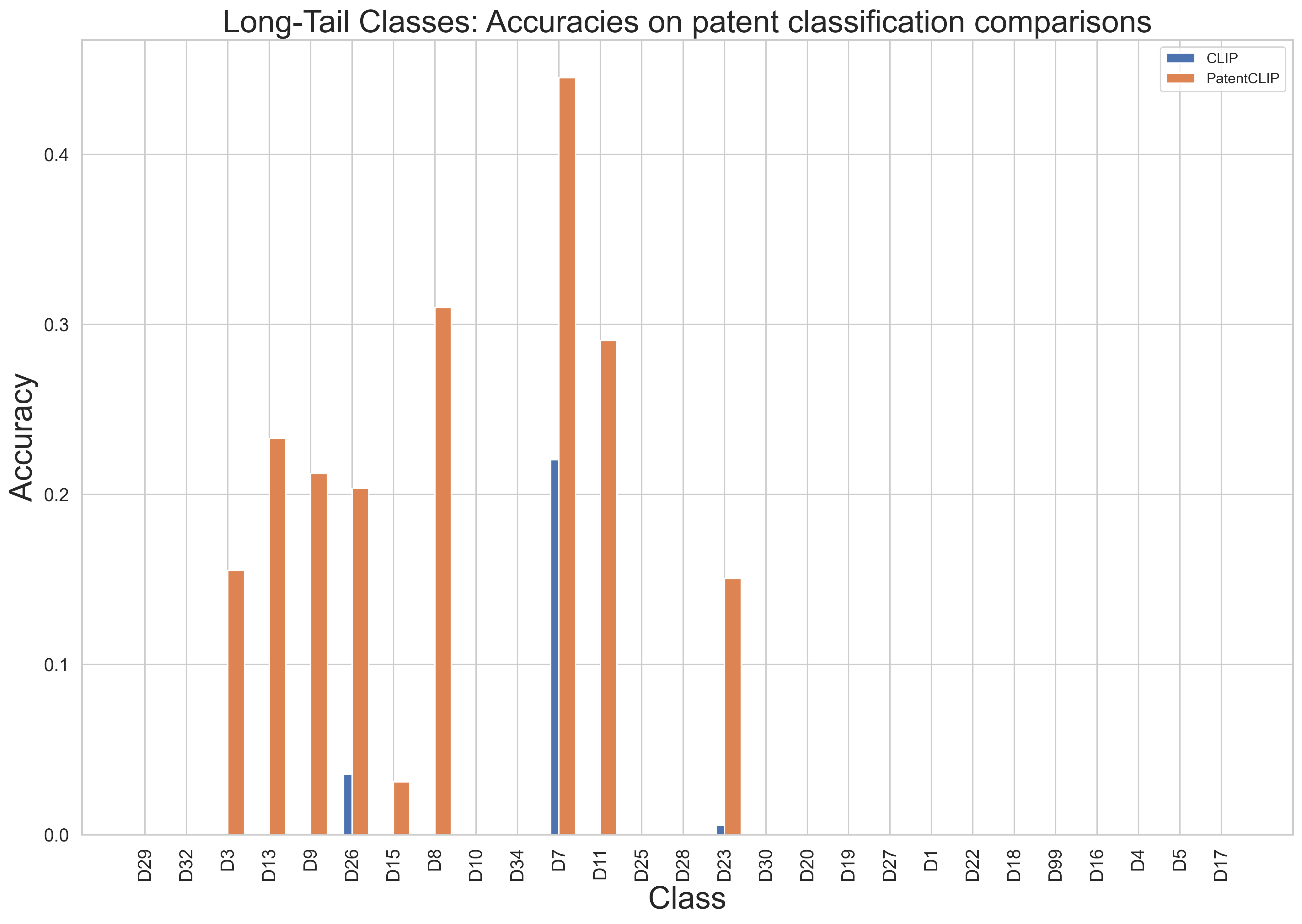}
         \caption{Patent classification comparisons on tail classes}
         \label{fig:rn101tail}      
     \end{subfigure}
        \caption{Comparisons between CLIP-RN101 and \ourmodel-RN101 in classification for top-6 classes and tail classes. \ourmodel improves significantly over CLIP on the all top classes. }
        \label{fig:confusion_scores}
\end{figure*}

\section{Additional Ablation Studies}
\label{app:ablation}

In this section, we illustrate additional ablation studies. We provide more results and analysis on the impact of different views. Moreover, we show two more ablation studies on the caption models for description generation and impact of vision backbones of \ourmodel. 

\subsection{Impact of different views}
\label{sec:diff_views}

Patent images feature views from various perspectives, and we explored how these views impact image retrieval. Table \ref{tab:multiviews} shows the retrieval performance across four different views with \ourmodel-ViT-B. The front view yields the highest recall, while the side and top views do not produce satisfactory results. We also combine these three views which we refer as `Multi view’ but this did not improve performance either. This lack of effectiveness can be attributed to the side and top views often not capturing sufficient design details. 
Based on these findings, we can conclude that the front view most effectively conveys the essential information of a design and aligns well with textual elements such as titles and captions.

\begin{table}[ht]
\centering

\begin{tabular}{cccccc}
\toprule
\multirow{2}{*}{View}  & \multicolumn{2}{c}{\bf{Text-Image}} & \multicolumn{2}{c}{\bf{Image-Text}} \\
  & R@5& R@10& R@5 & R@10  \\ \midrule

\multirow{1}{*}{Side view} &    8.24& 12.45&9.41& 13.61 \\
\multirow{1}{*}{Top view} &9.12  &13.37&9.96 &15.01 \\
\multirow{1}{*}{Multi view} &  8.59 &12.57 &9.58 &13.82 \\
\multirow{1}{*}{Front view}  &\bf{12.05}  &\bf{17.47} &\bf{13.36} &\bf{18.86} \\
\bottomrule
\end{tabular}
\caption{Multimodal Retrieval Task Performance using multiple views. Backbone is ViT-B. Front view demonstrated higher performance over the other three views. The highest Recall@K (\%) values are highlighted in bold.}
\label{tab:multiviews}
\end{table}

\subsection{Impact of vision backbones}
In training \ourmodel, we choose 4 backbone architectures RN50, RN101, ViT-B, and ViT-L. Table \ref{tab:patentclipfulldata} in the main paper shows  the performance of  text-image and image-text retrieval tasks for the comparisons of CLIP and our pretrained \ourmodel. The results show that both ResNet50 and ResNet101 achieve comparable recall, while ViT-B performs slightly better. Notably, the Vision Transformer Large (ViT-L) model significantly outperforms the others, as evidenced by its higher Recall@5 and Recall@10 scores in both retrieval directions (Text-Image and Image-Text). The size of the parameter of the ViT-L model allows it to better handle patent data and makes it effective for patent image and text retrieval tasks.

\section{More Qualitative Analysis}
In this section, we demonstrate  additional qualitative analysis. We begin with providing details of visualization of feature embeddings. Then, we show some multimodal retrieval examples. 

\subsection{More details of visualization of feature embeddings.}
Figure \ref{fig:feature} in the main paper compares UMAP \cite{mcinnes2018umap-software} visualizations of feature embeddings from CLIP and \ourmodel models for patent images. The patent data are from 2023, including 8,643 patents. We focus on the top most common categories: transportation, recording, lighting, games, furnishings and apparel. The CLIP model shows overlapping categories with less distinct clustering which indicates mixed feature recognition across various categories like transportation, games, and apparel. On the other hand, \ourmodel shows distinct clusters for almost all the categories. The clear clusters in the \ourmodel visualization show that it is very effective for accurately classifying and finding patent images.

\subsection{Additional class-aware classification analysis}
We present the performance comparisons of CLIP and \ourmodel with backbone of RN-101 on classification tasks. As results shown in Figure \ref{fig:rn101top}, \ourmodel improves significantly on the all top classes. Figure \ref{fig:rn101tail} shows that CLIP is not able to predict on most of the tail classes, but \ourmodel provides accurate predictions on 9 long-tail classes. Therefore, we believe that \ourmodel can improve the performance on patent classification and class-aware learning is effective for long-tail class distribution domain data.

\subsection{Image retrieval examples}
In this task, we present three examples of image-to-image retrieval using CLIP-ViT-B and \ourmodel-ViT-B as backbones with ArcFace \cite{deng2019arcface, higuchi2023patent} based on a given query image. The goal is to identify images similar to the query image. We show the top five retrieval results for both CLIP and \ourmodel for qualitative analysis. 

\begin{figure*}
        
     \centering
     \begin{subfigure}[b]{\textwidth}
         \centering
         \includegraphics[width=0.75\textwidth]{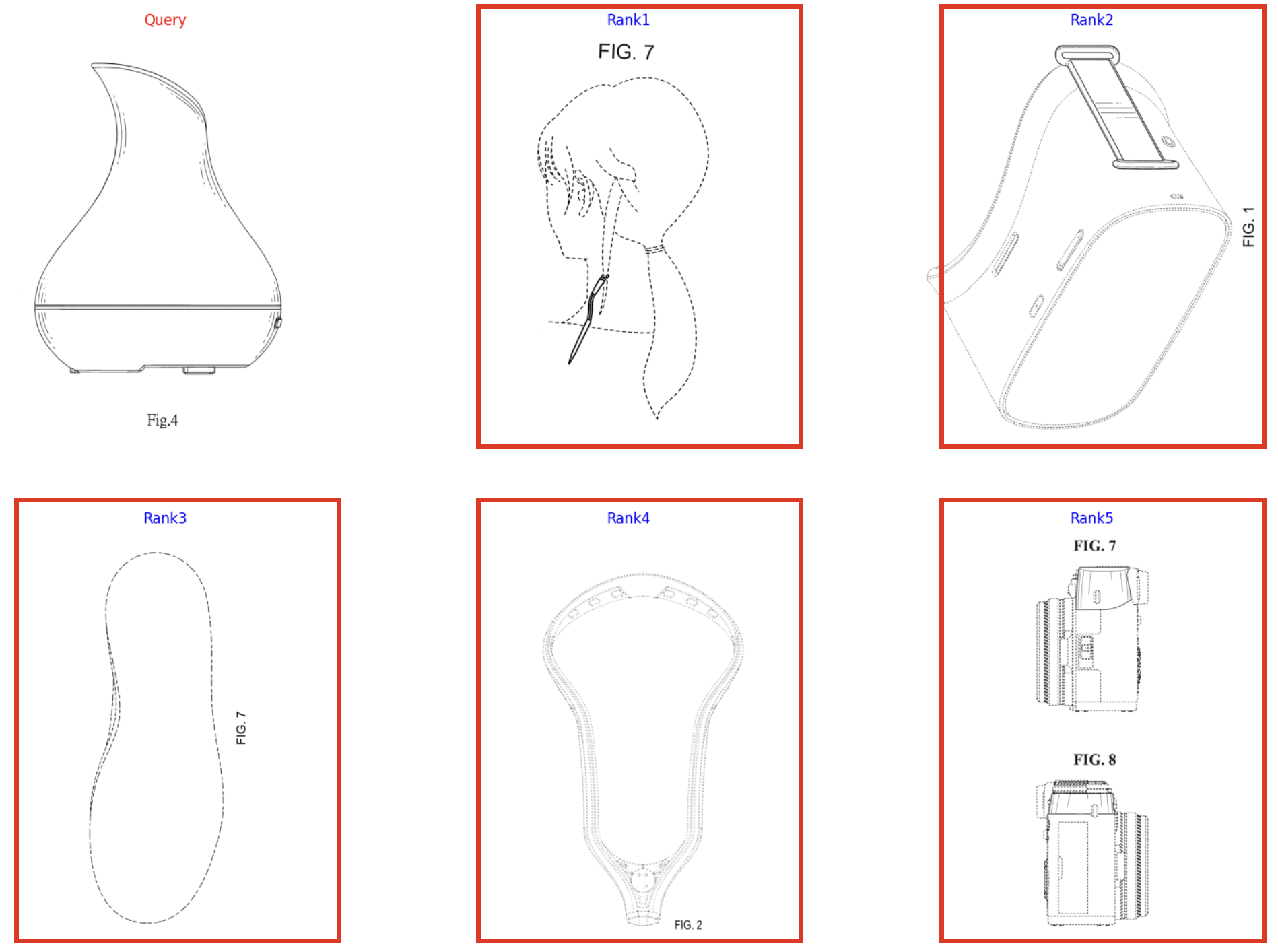}
         \caption{Image retrieval results of CLIP-ViT-B + ArcFace}
         \label{fig:irc1}
    \end{subfigure}
    \begin{subfigure}[b]{\textwidth}
         \centering
         \includegraphics[width=0.75\textwidth]{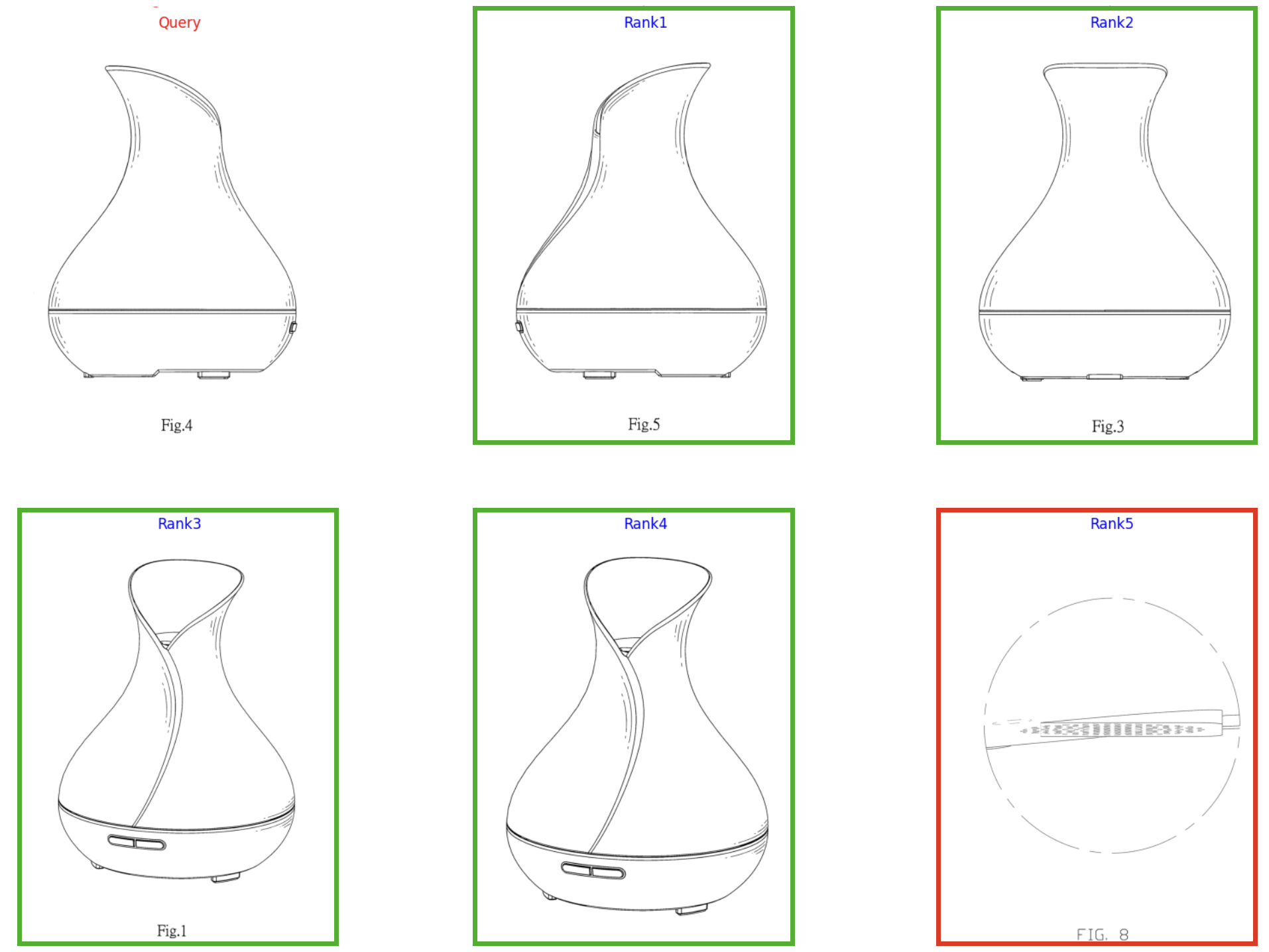}
         \caption{Image retrieval results of \ourmodel-ViT-B + ArcFace}
         \label{fig:irp1}
    \end{subfigure}

     \caption{Image Retrieval example 1 of patent D0815721. (a) and (b) are top 5 retrieval results of CLIP and \ourmodel respectfully. Green box denotes to the correct image. \ourmodel can retrieve the image correctly.}
     \label{fig:ir1}

\end{figure*}

\begin{figure*}
        
     \centering
     \begin{subfigure}[b]{\textwidth}
         \centering
         \includegraphics[width=0.75\textwidth]{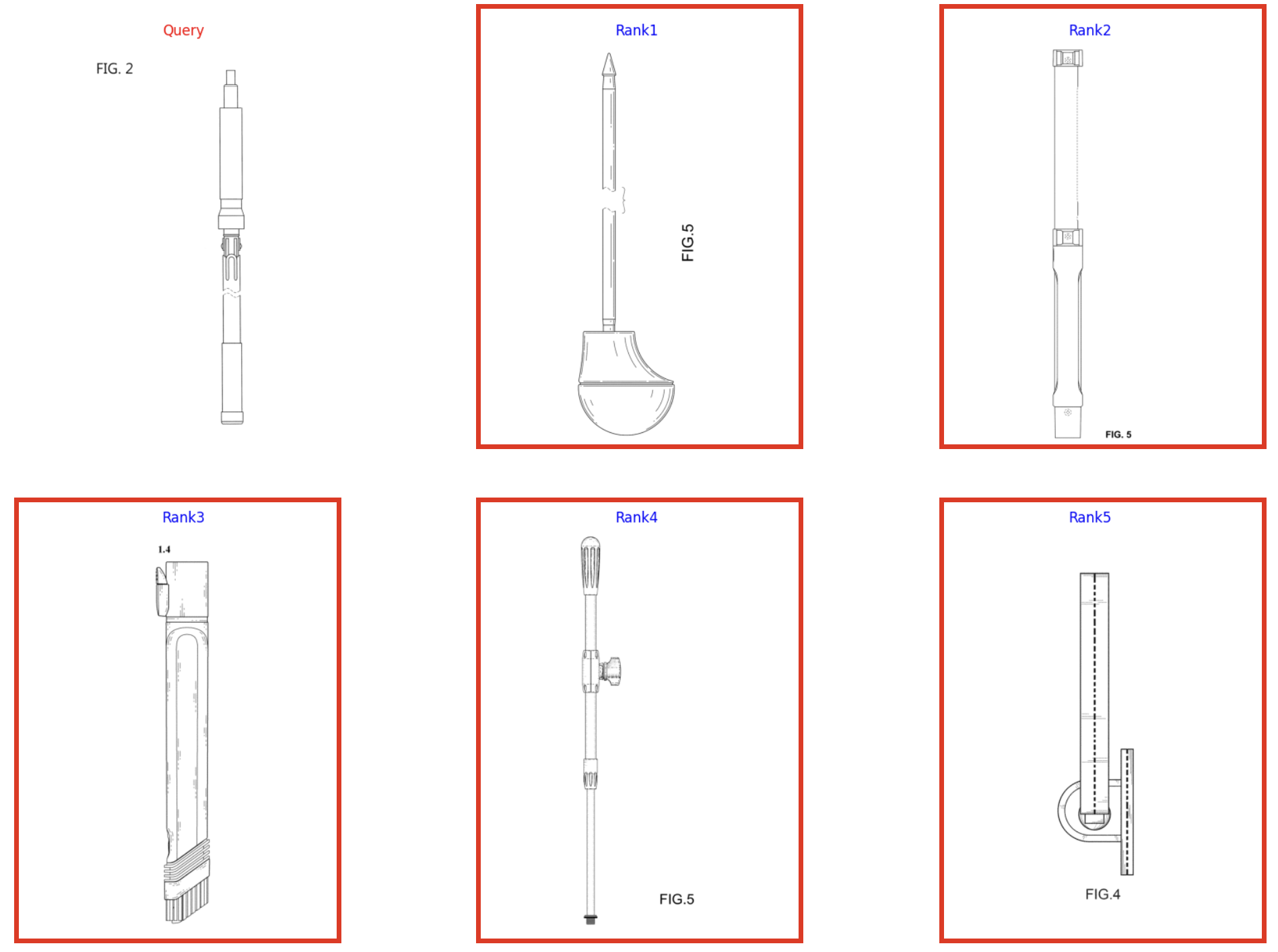}
         \caption{Image retrieval results of CLIP-ViT-B + ArcFace}
         \label{fig:irc2}
    \end{subfigure}
    \begin{subfigure}[b]{\textwidth}
         \centering
         \includegraphics[width=0.75\textwidth]{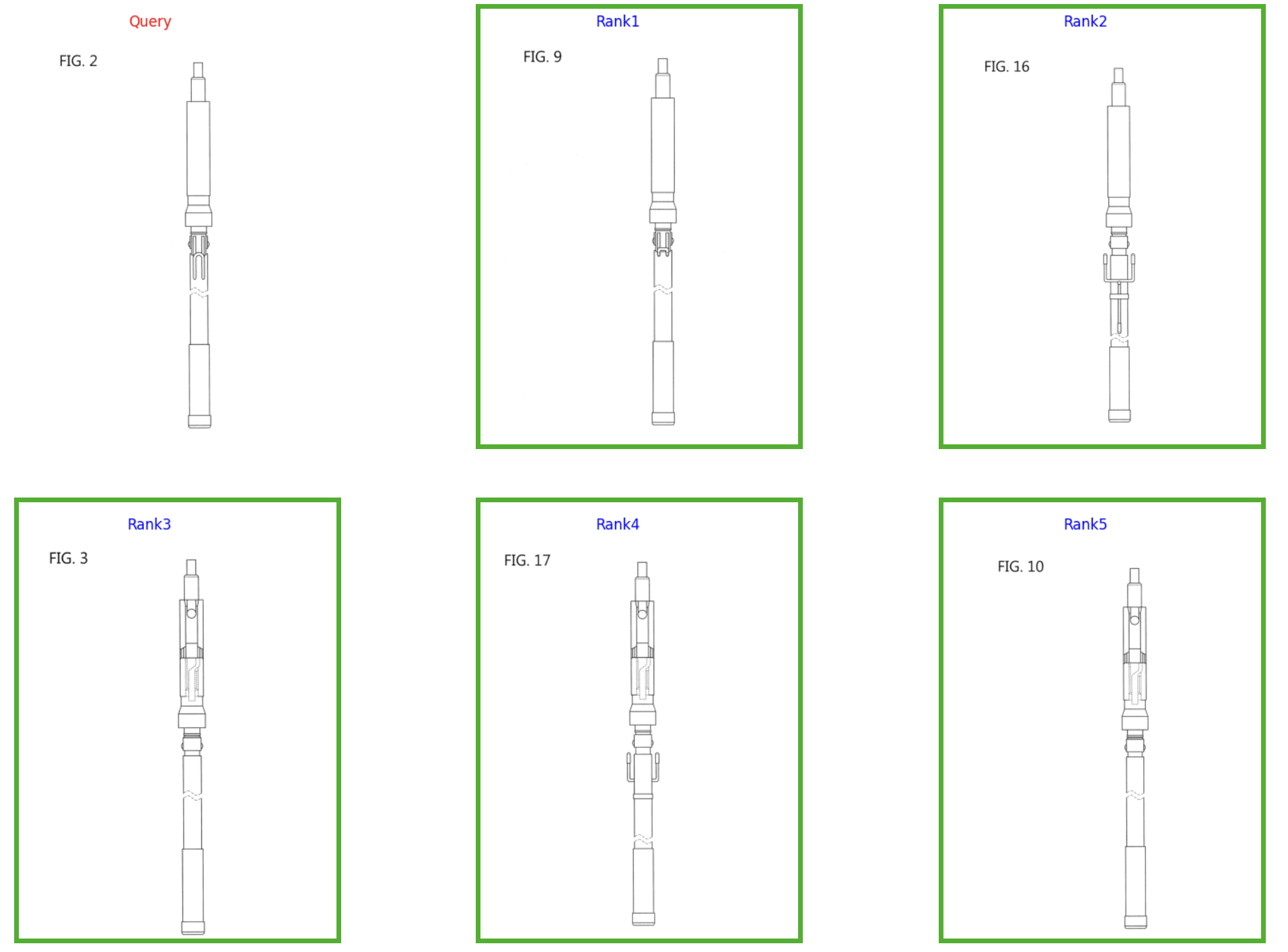}
         \caption{Image retrieval results of \ourmodel-ViT-B + ArcFace}
         \label{fig:irp2}
    \end{subfigure}

     \caption{Image Retrieval example 2 of patent D0817138. (a) and (b) are top 5 retrieval results of CLIP and \ourmodel respectfully. Green box denotes to the correct image. \ourmodel can retrieve the image correctly.}
     \label{fig:ir2}

\end{figure*}

\begin{figure*}
        
     \centering
     \begin{subfigure}[b]{\textwidth}
         \centering
         \includegraphics[width=0.75\textwidth]{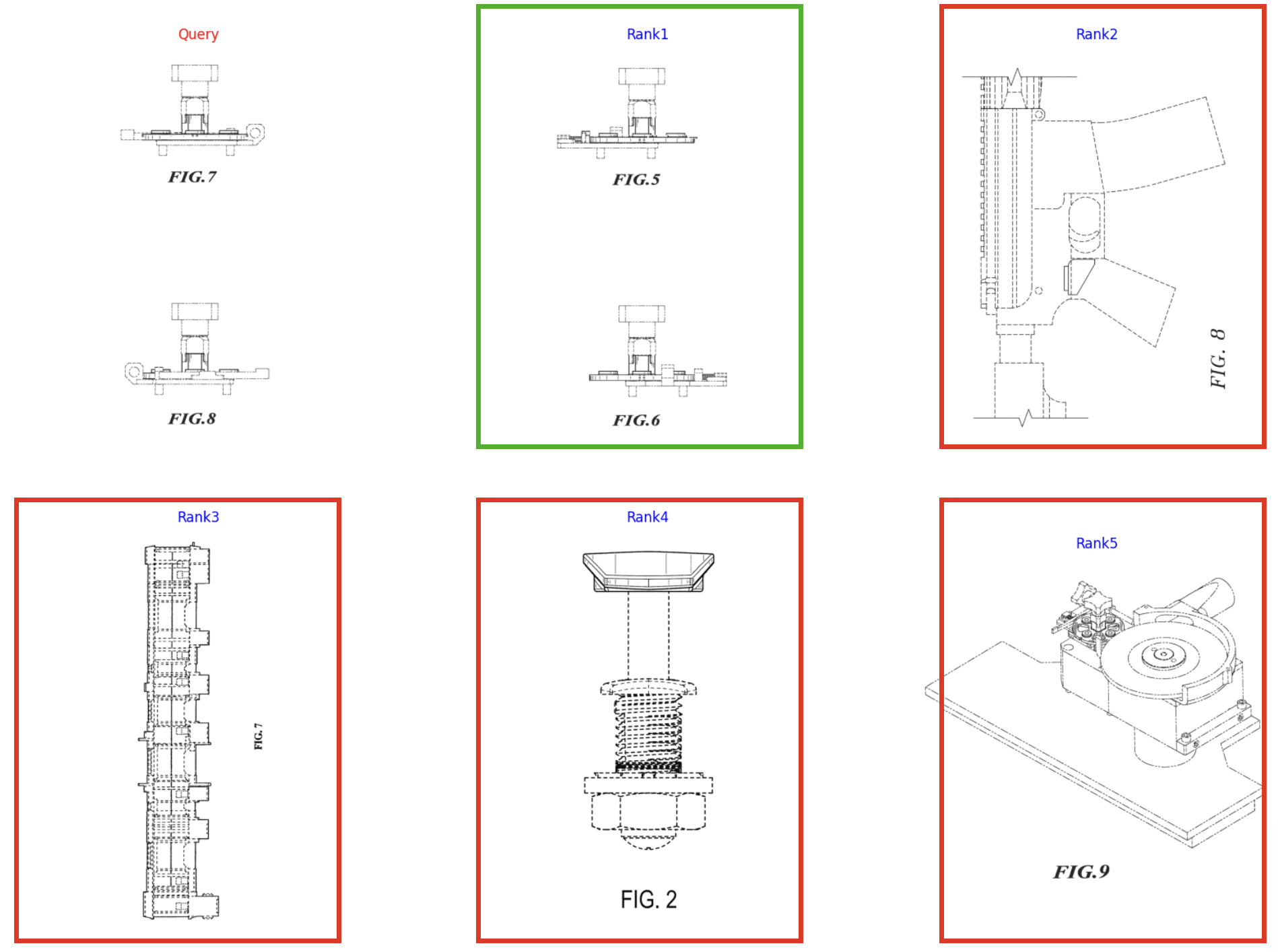}
         \caption{Image retrieval results of CLIP-ViT-B + ArcFace}
         \label{fig:irc3}
    \end{subfigure}
    \begin{subfigure}[b]{\textwidth}
         \centering
         \includegraphics[width=0.75\textwidth]{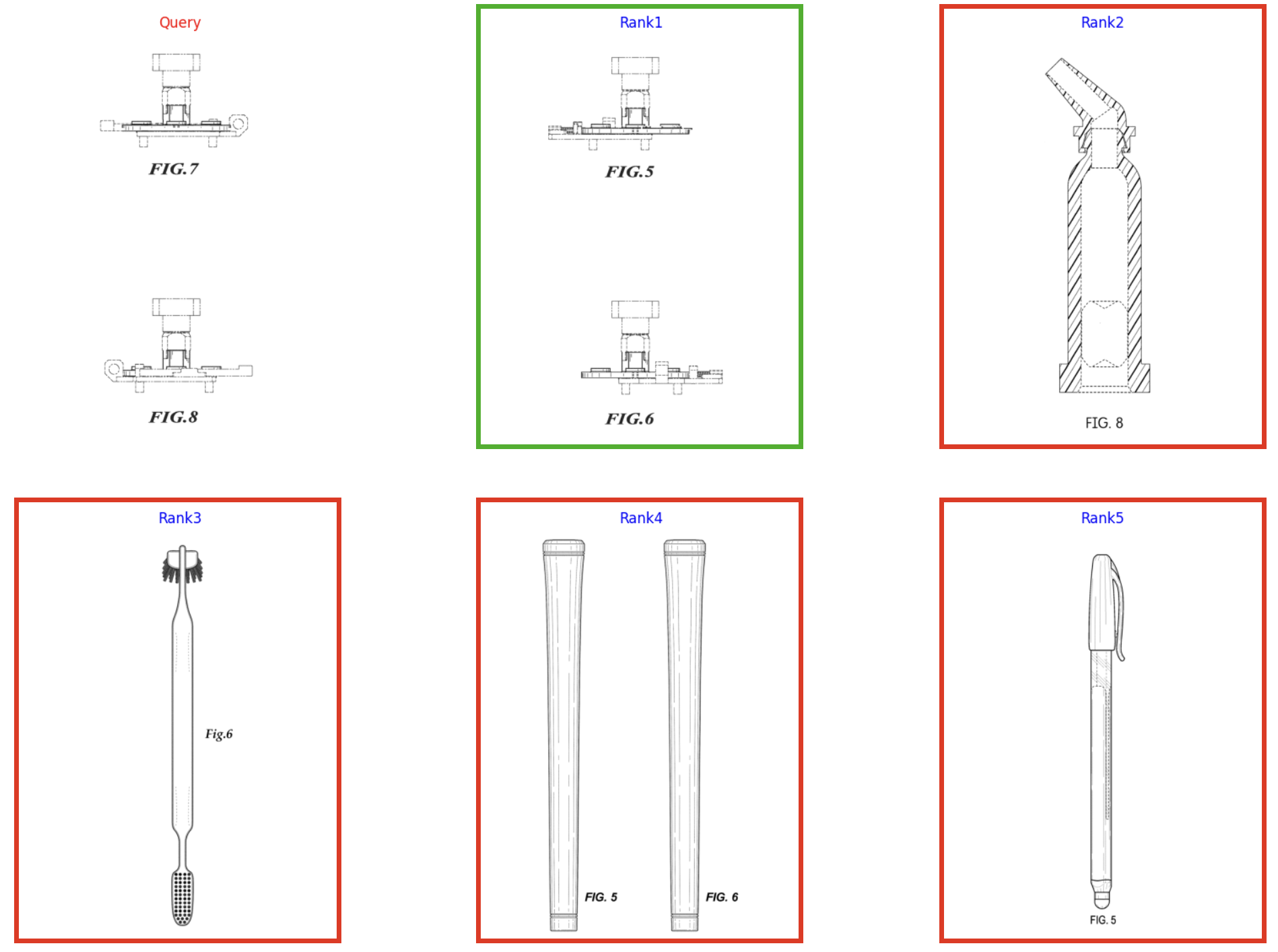}
         \caption{Image retrieval results of \ourmodel-ViT-B + ArcFace}
         \label{fig:irp3}
    \end{subfigure}

     \caption{Image Retrieval example 3 of patent id D0827684. (a) and (b) are top 5 retrieval results of CLIP and \ourmodel respectfully. Green box denotes to the correct image. CLIP and \ourmodel only retrieves rank 1 image correctly.}
     \label{fig:ir3}

\end{figure*}

In Example 1, shown in Figure \ref{fig:ir1}, CLIP fails to retrieve a single relevant image, while \ourmodel correctly retrieves the top 4 out of 5 images. In Figure \ref{fig:ir2}, our model achieves perfect retrieval, identifying all 5 relevant images in different views. We observe that CLIP only can retrieve images with similar shape but fails to capture patent class and different views. Therefore, \ourmodel pre-trained with class-aware information and multi-views can improve performance in image retrieval tasks. However, in Figure \ref{fig:ir3}, both CLIP and our model demonstrate similar performance. We leave the analysis of unsuccessful cases---such as Figure \ref{fig:ir3}---for our future work.




\subsection{Multimodal retrieval examples.}We present four examples of multimodal image retrieval. For each example, based on a given query (caption) and a ground truth (GT) image, the objective is to identify similar images to the GT. The backbone is ViT-B. We show the top five retrieval results for both the CLIP model and \ourmodel for case studies of their performance comparisons in matching images to textual descriptions. 

\begin{itemize}
    \item Example 1: \textit{Text Query:} The image is a drawing of a wheel, which is a circular object with a central hub and spokes.
    
    \textit{Ground truth image:} D0862341.TIF, where D0862341 is patent id.
    
    \item Example 2: \textit{Text Query:} The image is a drawing of a control valve. The control valve is a device used to regulate the flow of fluid, such as water, steam, or gas, in a system.
    
    \textit{Ground truth image:} D0858713.TIF
    
    \item Example 3: \textit{Text Query:} The image is a truck vehicle grille. The grille serves as a protective covering for the front of the truck, covering the engine and radiator.
    
    \textit{Ground truth image:} D0850330.TIF

     \item Example 3: \textit{Text Query:} The image is a square-shaped vehicle floor mat, which is designed to provide comfort and protection for the vehicle floor.
    
    \textit{Ground truth image:} D0845852.TIF
\end{itemize}

\begin{figure*}
        
     \centering
     \begin{subfigure}[b]{\textwidth}
     \centering
         \includegraphics[width=0.85\textwidth]{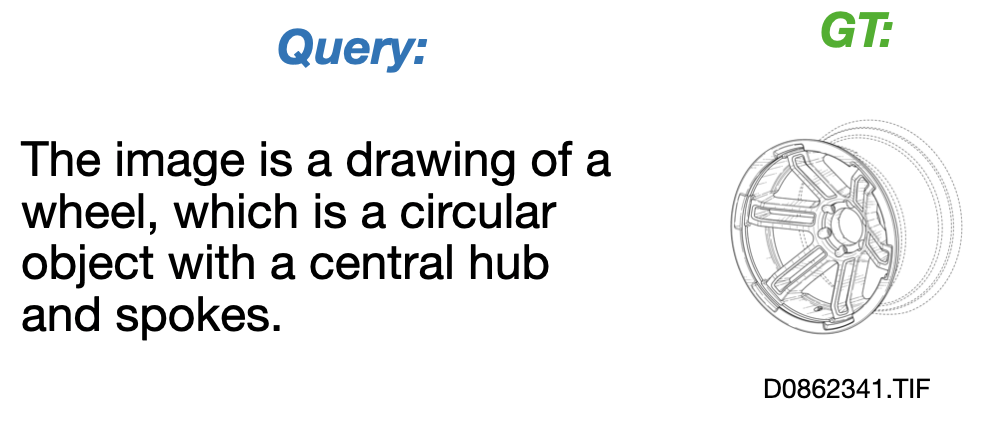}
         \caption{Text query and ground truth image}
         \label{fig:gt1}
    \end{subfigure}
    \begin{subfigure}[b]{\textwidth}
    \centering
         \includegraphics[width=.85\textwidth]{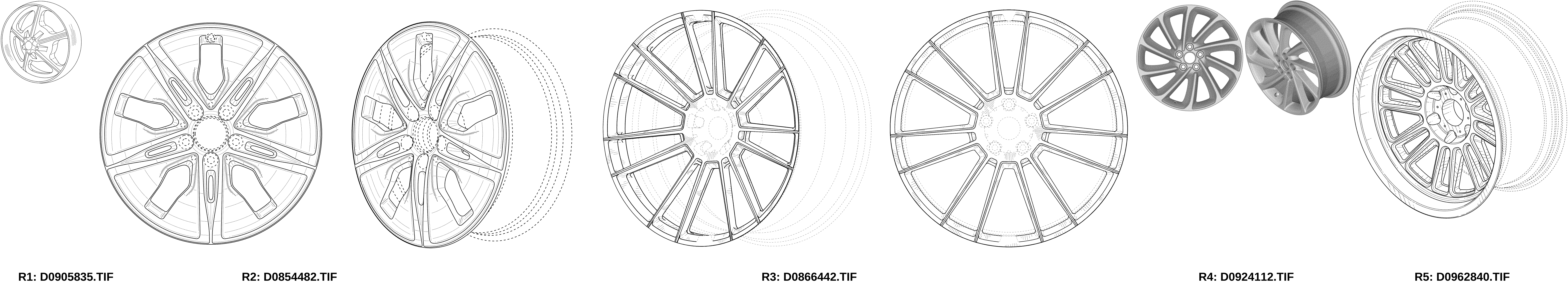}
         \caption{Retrieval results of CLIP}
         \label{fig:cr1}
    \end{subfigure}
    \begin{subfigure}[b]{\textwidth}
    \centering
         \includegraphics[width=.85\textwidth]{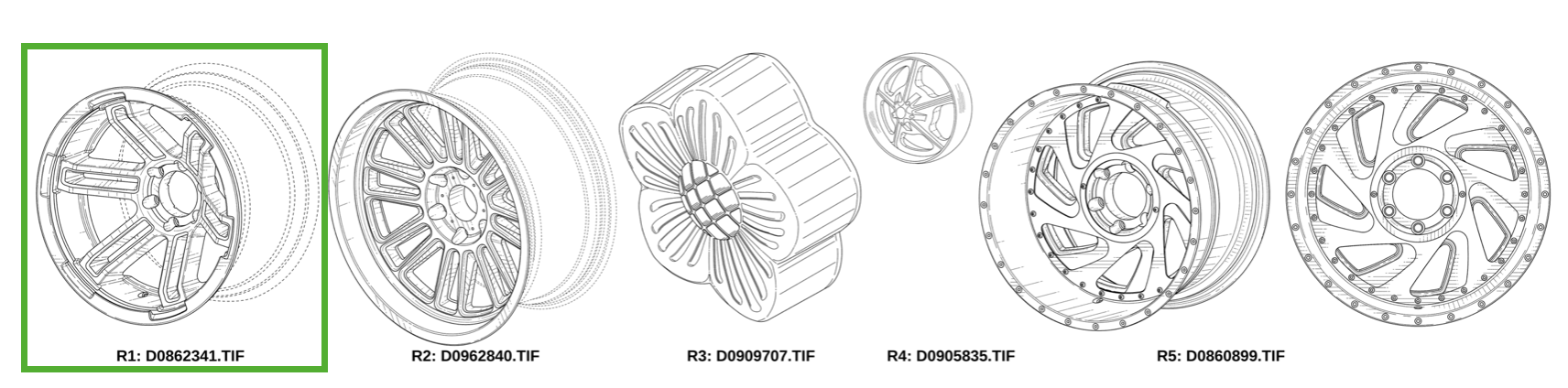}
         \caption{Retrieval results of \ourmodel}
         \label{fig:pcr1}
     \end{subfigure}

     \caption{Text-image Retrieval example 1. Text query and ground truth image are shown in (a). (b) and (c) are top 5 retrieval results of CLIP and \ourmodel respectfully. Top 1-5 is from left to right. Green box denotes to the correct image. \ourmodel can retrieve the image correctly.}
     \label{fig:r1}

\end{figure*}

\begin{figure*}
        
     \centering
     \begin{subfigure}[b]{\textwidth}
         \centering
         \includegraphics[width=0.8\textwidth]{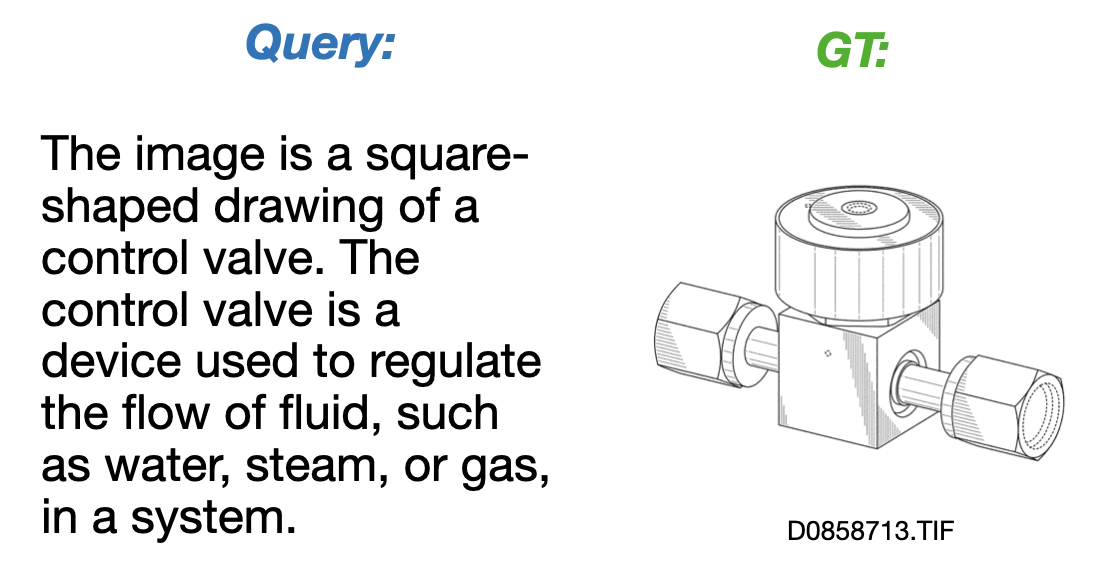}
         \caption{Text query and ground truth image}
         \label{fig:gt2}
    \end{subfigure}
    \begin{subfigure}[b]{\textwidth}
         \centering
         \includegraphics[width=0.8\textwidth]{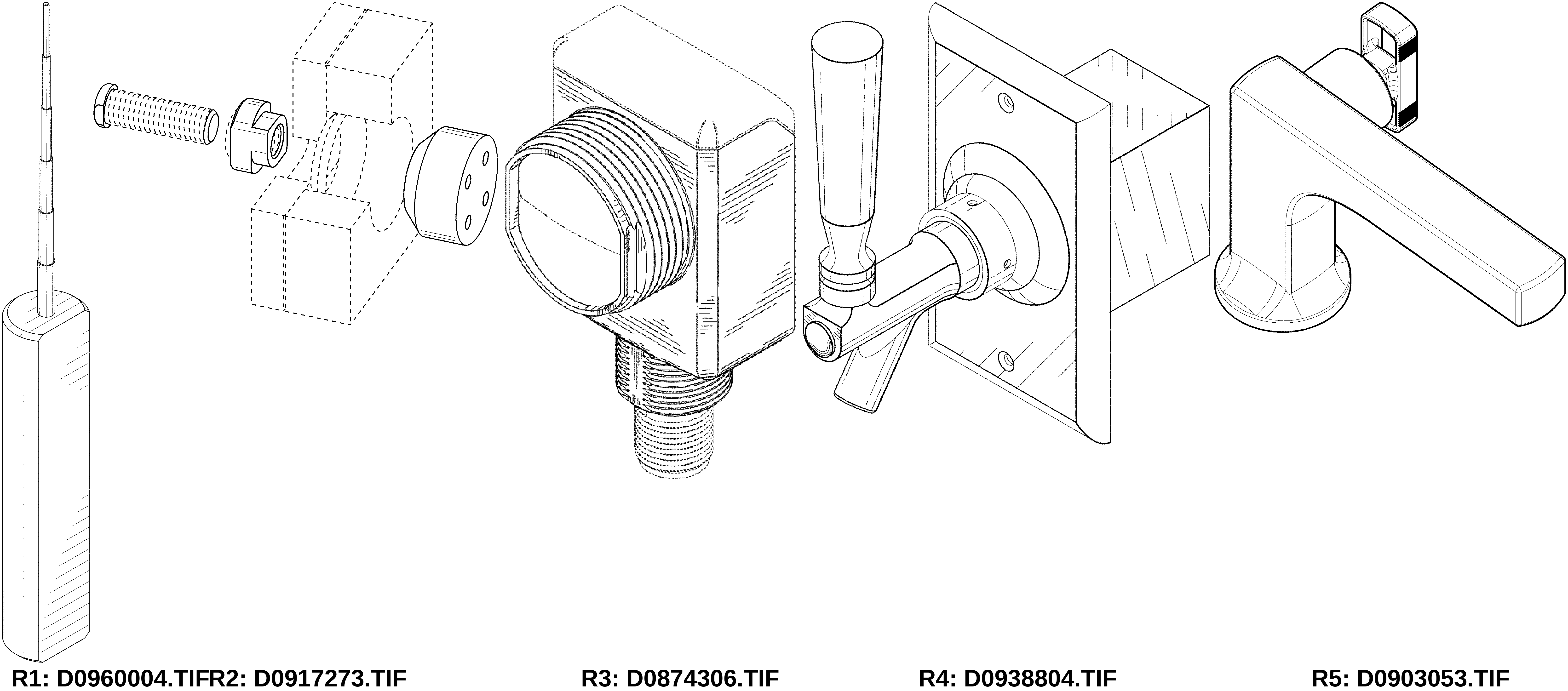}
         \caption{Retrieval results of CLIP}
         \label{fig:cr2}
    \end{subfigure}
    \begin{subfigure}[b]{\textwidth}
    \centering
         \includegraphics[width=0.95\textwidth]{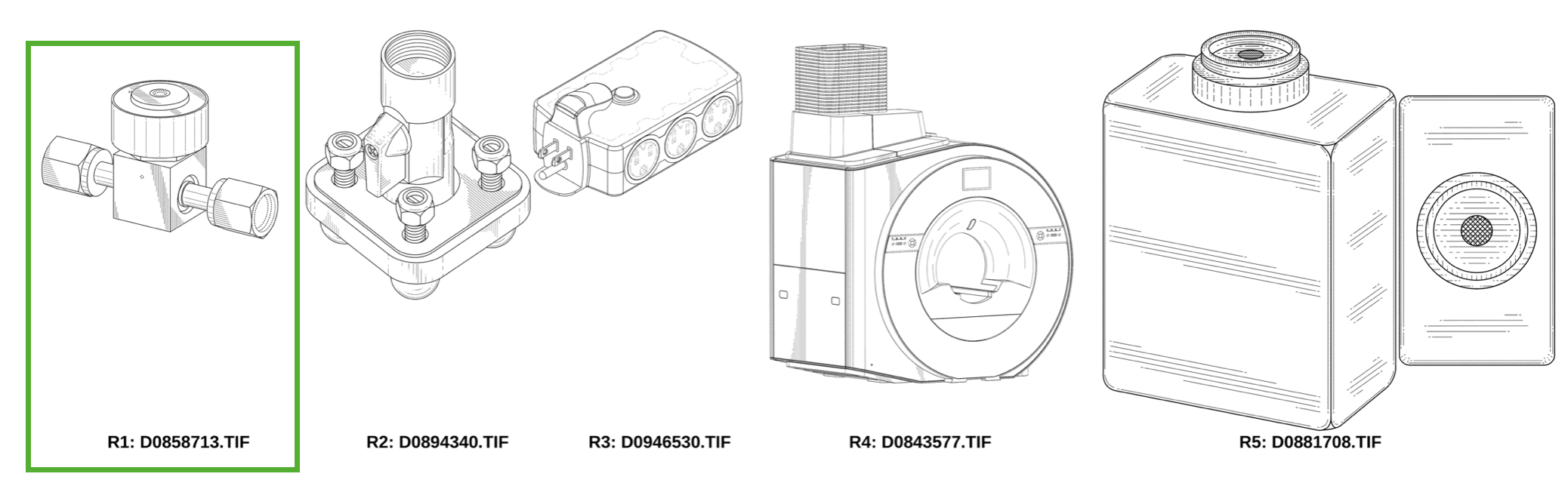}
         \caption{Retrieval results of \ourmodel}
         \label{fig:pcr2}
     \end{subfigure}

     \caption{Text-image Retrieval example 2. Text query and ground truth image are shown in (a). (b) and (c) are top 5 retrieval results of CLIP and \ourmodel respectfully. Top 1-5 is from left to right. Green box denotes to the correct image. \ourmodel can retrieve the image correctly.}
     \label{fig:r2}

\end{figure*}

\begin{figure*}
        
     \centering
     \begin{subfigure}[b]{\textwidth}
     \centering
         \includegraphics[width=0.9\textwidth]{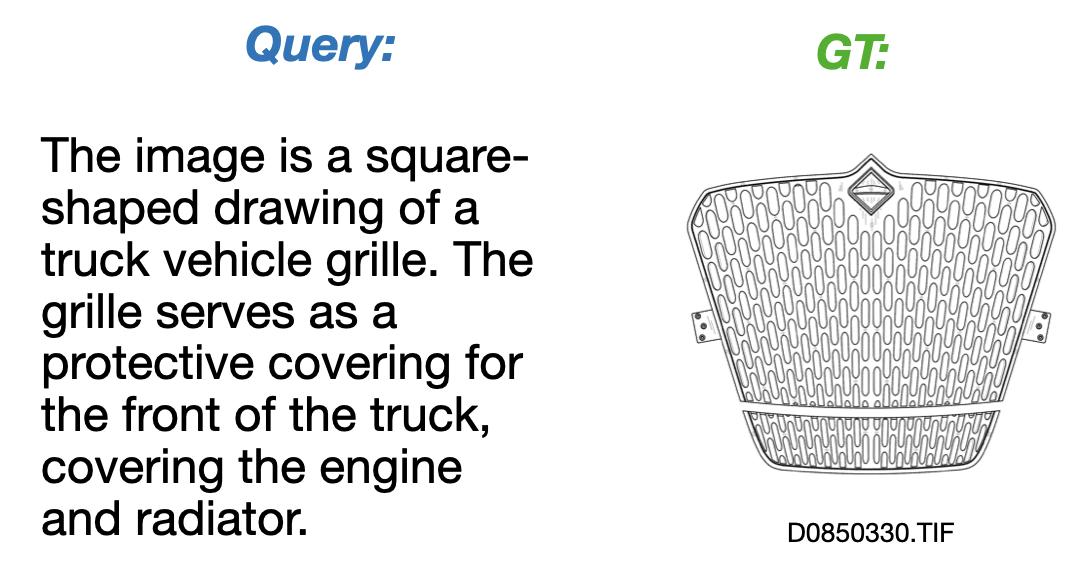}
         \caption{Text query and ground truth image}
         \label{fig:gt3}
    \end{subfigure}
    \begin{subfigure}[b]{\textwidth}
    \centering
         \includegraphics[width=0.9\textwidth]{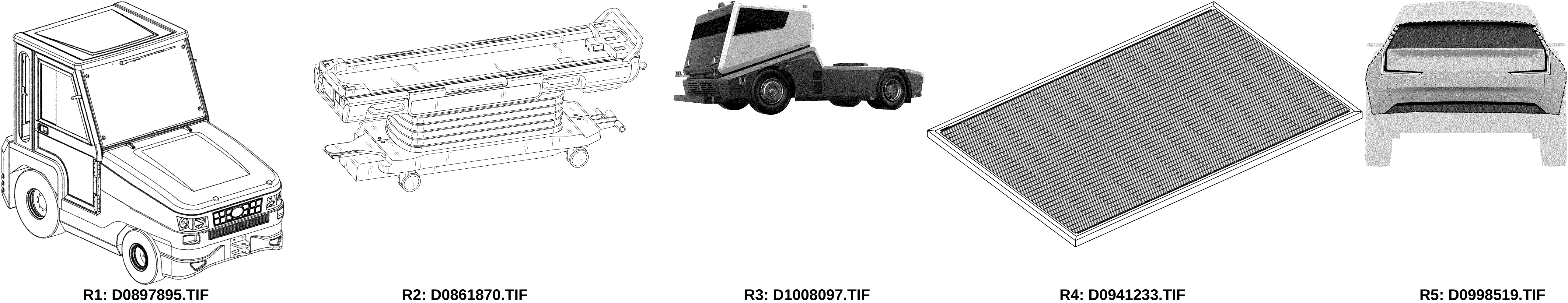}
         \caption{Retrieval results of CLIP}
         \label{fig:cr3}
    \end{subfigure}
    \begin{subfigure}[b]{\textwidth}
    \centering
         \includegraphics[width=0.9\textwidth]{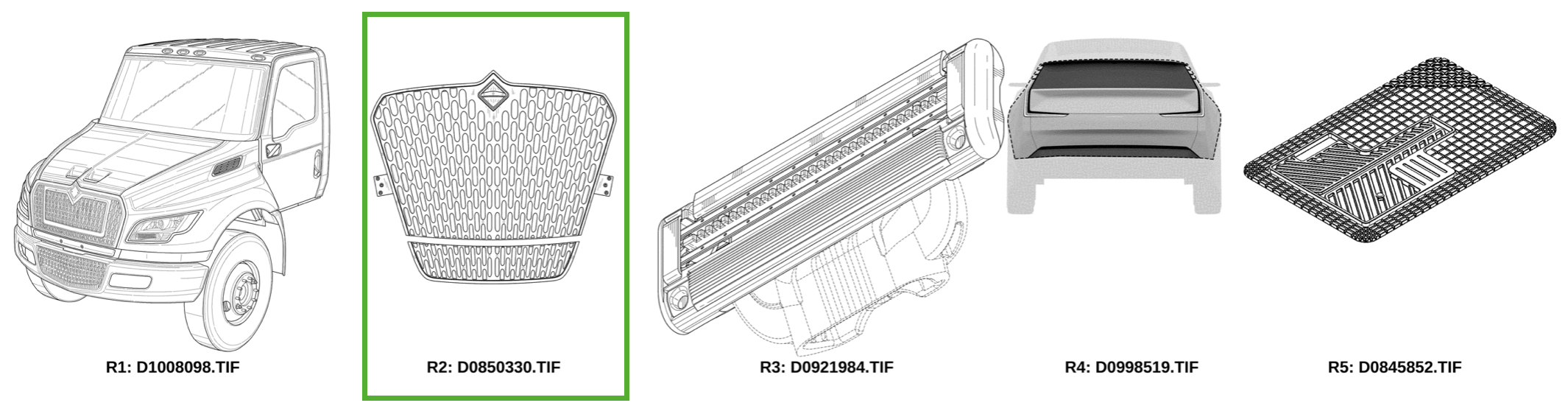}
         \caption{Retrieval results of \ourmodel}
         \label{fig:pcr3}
     \end{subfigure}

     \caption{Text-image Retrieval example 3. Text query and ground truth image are shown in (a). (b) and (c) are top 5 retrieval results of CLIP and \ourmodel respectfully. Top 1-5 is from left to right. Green box denotes to the correct image. \ourmodel can retrieve the image correctly.}
     \label{fig:r3}

\end{figure*}

\begin{figure*}
        
     \centering
     \begin{subfigure}[b]{\textwidth}
         \centering
         \includegraphics[width=0.9\textwidth]{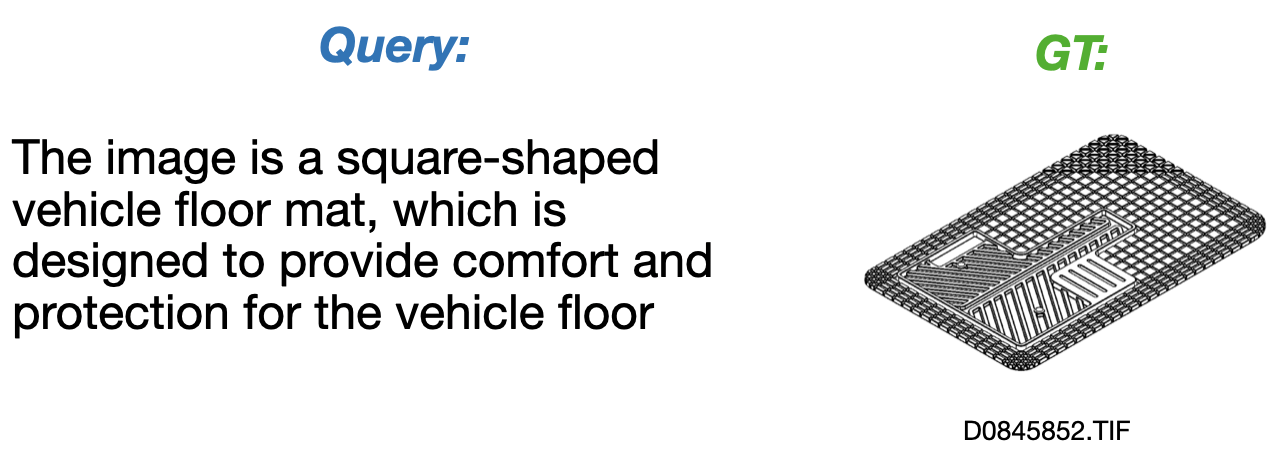}
         \caption{Text query and ground truth image}
         \label{fig:gt4}
    \end{subfigure}
    \begin{subfigure}[b]{\textwidth}
    \centering
         \includegraphics[width=.9\textwidth]{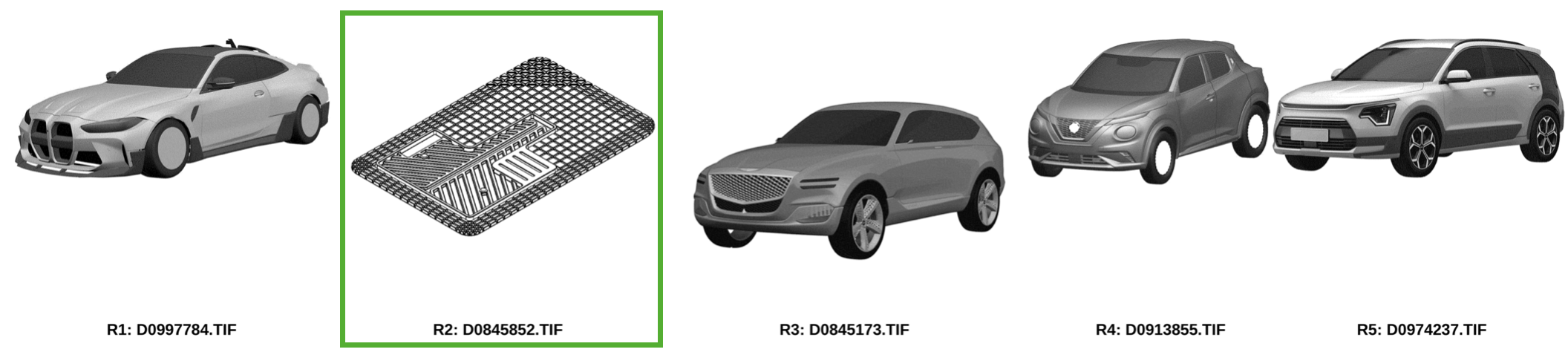}
         \caption{Retrieval results of CLIP}
         \label{fig:cr4}
    \end{subfigure}
    \begin{subfigure}[b]{\textwidth}
    \centering
         \includegraphics[width=.9\textwidth]{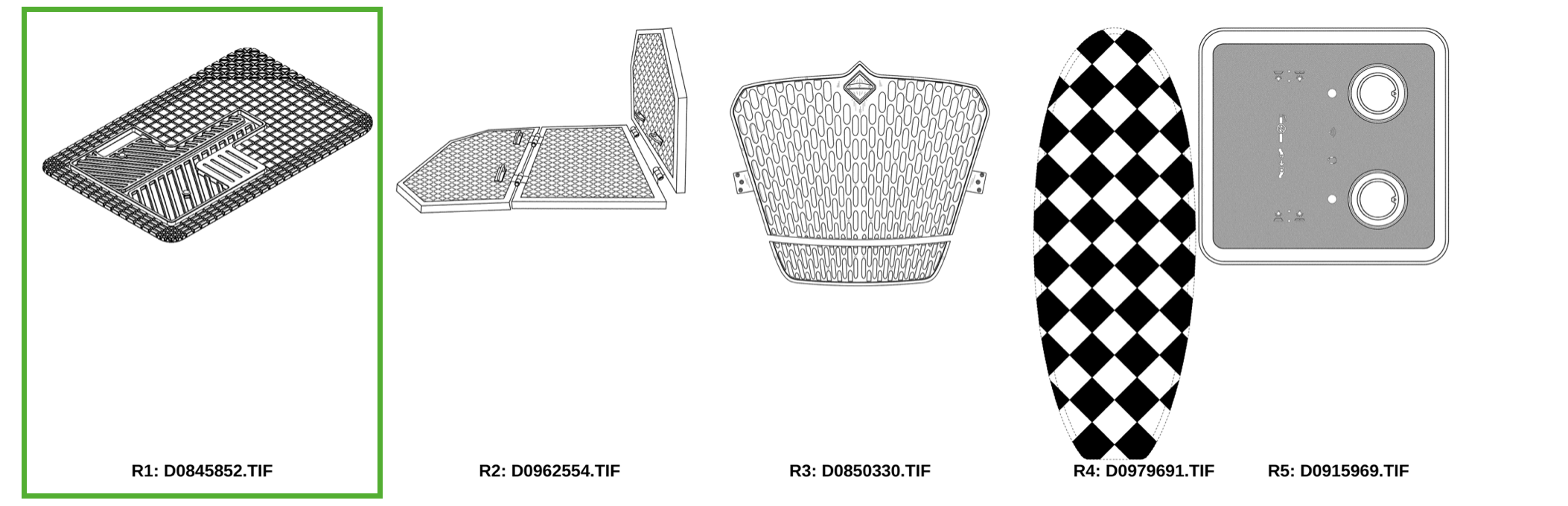}
         \caption{Retrieval results of \ourmodel}
         \label{fig:pcr4}
     \end{subfigure}

     \caption{Text-image Retrieval example 4. Text query and ground truth image are shown in (a). (b) and (c) are top 5 retrieval results of CLIP and \ourmodel respectfully. Top 1-5 is from left to right. Green box denotes to the correct image. Both \ourmodel and CLIP can retrieval the ground truth image, but \ourmodel retrieve top 1 image correctly.}
     \label{fig:r4}

\end{figure*}

\textbf{Our \ourmodel model retrieves successfully the ground truth images as the top results in Figures \ref{fig:r1},  \ref{fig:r2}, and \ref{fig:r4}}. In \ref{fig:r3}, the ground truth image was retrieved as the second top result. Specifically, as shown in Figure \ref{fig:r1}, both models can retrieve wheels. However, \ourmodel is pre-trained on our proposed methods tailored on the design patent data, it is able to capture the details of design which demonstrates \ourmodel can provide better capability for prior art search and design inspirations. Moreover, in \ref{fig:r4}, although CLIP retrieved ground truth images, other retrieved images are vehicles, not a floor mat. However, \ourmodel can retrieve images that are more relevant for vehicle mats. In fact, \ourmodel learns better representations for understanding design patent data in multimodal scenarios.

\end{document}